\documentclass{article} 
\usepackage{iclr2023_preprint,times}

\usepackage{graphicx}
\usepackage{amssymb,amsmath}
\usepackage{hyperref}
\usepackage{url}
\usepackage{xcolor}

\usepackage{algorithm}
\usepackage{algpseudocode}
\usepackage{tabto}

\title{
Reward Learning with Trees:\\Methods and Evaluation
}

\author{Tom Bewley$^1$, Jonathan Lawry$^1$, Arthur Richards$^1$, Rachel Craddock$^2$ \& Ian Henderson$^2$ \\
$^1$University of Bristol and $^2$Thales, United Kingdom \\
\texttt{\{firstname.surname\}@\{}$^1$\texttt{bristol.ac.uk,}$^1$\texttt{uk.thalesgroup.com\}} \\
}

\newcommand{\sdot}{,\hspace{-0.04cm}.\hspace{-0.01cm}.\hspace{-0.01cm}.,}

\iclrfinalcopy 
\begin{document}

\maketitle

\vspace{-0.525cm}
\begin{abstract}
\vspace{-0.15cm}
Recent efforts to learn reward functions from human feedback have tended to use deep neural networks, whose lack of transparency hampers our ability to explain agent behaviour or verify alignment. We
explore the merits of
learning intrinsically interpretable tree models instead. We develop a recently proposed method for learning reward trees from preference labels, and show it to be broadly competitive with neural networks on challenging high-dimensional tasks, with good robustness to limited or corrupted data. Having found that reward tree learning \textit{can} be done effectively in complex settings, we then consider why it \textit{should} be used, demonstrating that the interpretable reward structure gives significant scope for traceability, verification and explanation.
\end{abstract}

\vspace{-0.25cm}
\section{Introduction}

\vspace{-0.15cm}
For a reinforcement learning (RL) agent to reliably achieve a goal or desired behaviour, this objective must be encoded as a reward function. However, manual reward design is widely understood to be challenging, with risks of under-, over-, and mis-specification leading to undesirable, unsafe and variable outcomes \citep{pan2022the}. For this reason, there has been growing interest in enabling RL agents to learn reward functions from normative feedback provided by humans
\citep{leike2018scalable}.
These efforts have proven successful from a technical perspective, but an oft-unquestioned aspect of the approach creates a roadblock to practical applications: reward learning typically uses black-box neural networks (NNs), which resist human scrutiny and interpretation.

\vspace{-0.13cm}
For advocates of explainable AI (XAI), this is a problematic state of affairs. The XAI community is vocal about the safety and accountability risks of opaque learning algorithms \citep{rudin2019stop}, but an inability to interpret even the objective that an agent is optimising places us in yet murkier epistemic territory, in which an understanding of the causal origins of learnt behaviour, and their alignment with human preferences, becomes virtually unattainable.
Black-box reward learning could also be seen as a missed scientific opportunity. A learnt reward function is a tantalising object of study from an XAI perspective, due to its triple status as (1) an \textit{explanatory}
model of revealed human preferences, (2) a \textit{normative} model of agent behaviour, and (3) a \textit{causal} link between the two.

\vspace{-0.13cm}
The approach proposed by \citet{bewley2022interpretable} provides a promising way forward. Here, human preference labels over pairs of agent behaviours are used to learn tree-structured reward functions (\textit{reward trees}), which are hierarchies of local rules that admit visual and textual representation and can be leveraged to monitor and debug agent learning. In this paper, we adapt and extend the method (including by integrating it with model-based RL agents), and
compare it to
NN-based reward learning in a challenging aircraft handling domain.
We find it to be broadly competitive on both quantitative metrics and qualitative assessments, with our new modification to tree growth yielding significant improvements. The resultant trees are small enough to be globally interpretable ($\approx 20$ leaves), and we demonstrate how they can be analysed, verified, and used to generate explanations.

\vspace{-0.13cm}
The primary contribution of this paper is positive evidence that
reward learning can be performed effectively using interpretable models such as trees, even in complex, high-dimensional continuous environments. Our secondary contributions are improvements to the originally-proposed learning algorithm, as well as metrics and methods for reward evaluation and interpretability that may be useful to others working in what remains a somewhat preparadigmatic field. After reviewing the necessary background and related work in Sections \ref{sec:related_work} and \ref{sec:pbrl}, we present our refinement of reward tree learning in Section \ref{sec:model_induction}, and describe how we deploy it online with a model-based agent in Section \ref{sec:online_learning}. Section \ref{sec:results} contains our experiments and results, which consider both quantitative and qualitative aspects of learning performance, and an illustrative analysis of learnt tree structures. Finally, Section \ref{sec:conclusion} concludes and discusses avenues for future work.

\section{Background and Related Work}
\label{sec:related_work}

\vspace{-0.1cm}
\paragraph{Markov Decision Processes (MDPs)} In this canonical formulation of sequential decision making, the state of a system at discrete time $t$, $s_t\in\mathcal{S}$, and the action of an agent, $a_t\in\mathcal{A}$, condition the successor state $s_{t+1}$ according to a dynamics function $D:\mathcal{S}\times\mathcal{A}\rightarrow\Delta(\mathcal{S})$ (we use $\Delta(\cdot)$ to denote the set of all probability distributions over a set). A reward function $R:\mathcal{S}\times\mathcal{A}\times\mathcal{S}\rightarrow\mathbb{R}$ then outputs a scalar reward $r_{t+1}$ given $s_t$, $a_t$ and $s_{t+1}$. RL algorithms use exploratory data collection to learn action-selection policies $\pi:\mathcal{S}\rightarrow\Delta(\mathcal{A})$, with the goal of maximising the expected discounted sum of future reward $\mathbb{E}_{D,\pi}\sum_{h=0}^\infty\gamma^h r_{t+h+1},\gamma\in[0,1]$.

\vspace{-0.3cm}
\paragraph{Reward Learning} In the usual MDP framing, $R$ is an immutable property of the environment, which belies the practical fact that AI objectives originate in the uncertain goals and preferences of fallible humans \citep{russell2019human}. Reward learning (or modelling) \citep{leike2018scalable} replaces hand-specified reward functions with models learnt from humans via revealed preference cues
such as demonstrations \citep{ng2000algorithms}, scalar evaluations \citep{knox2008tamer}, approval labels \citep{griffith2013policy}, corrections \citep{bajcsy2017learning}, and preference rankings \citep{christiano2017deep}.

\vspace{-0.3cm}
\paragraph{XAI for RL (XRL)}
Surveys of XAI for RL \citep{puiutta2020explainable,heuillet2021explainability} taxonomise a diverse and expanding range of methods. A key division is between intrinsic approaches, which imbue agents with structure such as object-oriented representations \citep{zhu2018object} or symbolic policy primitives \citep{verma2018programmatically}, and post hoc (often visual) analyses of learnt representations \citep{zahavy2016graying}, including computing feature importance/saliency \citep{huber2019enhancing}. Spatiotemporal scope varies from the local explanation of single actions \citep{van2018contrastive} to the global summary of entire policies by showing representative trajectories \citep{amir2018highlights} or critical states \citep{huang2018establishing}. While most post hoc methods focus on single policies, some provide insight into the dynamics of agent learning \citep{dao2018deep,bewley2022summarising}.

\vspace{-0.3cm}
\paragraph{Explainable Reward Functions}
At the intersection of reward learning and XRL lie efforts to improve human understanding of reward functions and their effects on action selection. While this area is \textit{``less developed"} than other XRL sub-fields \citep{glanois2021survey}, a distinction has again emerged between intrinsic approaches which create rewards that decompose into semantic components \citep{juozapaitis2019explainable} or optimise for sparsity \citep{devidze2021explicable}, and post hoc approaches which apply feature importance analysis \citep{russell2019explaining}, counterfactual probing \citep{michaud2020understanding}, or simplifying transformations \citep{jenner2022preprocessing}. \citet{sanneman2022empirical} use a set of human-oriented metrics to compare the efficacy of reward explanation techniques. 

\vspace{-0.3cm}
\paragraph{Trees for Explainable Agency}
Prior uses of tree models in XRL again divide into intrinsic methods, in which an agent's policy \citep{pmlr-v108-silva20a}, value function \citep{liu2018toward,roth2019conservative} or dynamics model \citep{jiang2019experience} is implemented as a tree, and post hoc tree approximations of an existing agent's (usually NN) policy \citep{bastani2018verifiable,coppens2019distilling} or transitions in the environment \citep{bewley2022summarising}. Related to our focus on human-centric learning: \citet{cobo2012automatic} learn tree-structured MDP abstractions from human demonstrations; \citet{lafond2013cognitive} use tree models to model expert judgements in a naval air defence setting; and \citet{tambwekar2021specifying} warm-start RL by converting a natural language specification into a differentiable tree policy.

\vspace{-0.2cm}
\section{Preference-based Reward Learning}
\label{sec:pbrl}

\vspace{-0.1cm}
We adopt the preference-based approach to reward learning, in which a human is presented with pairs of agent trajectories (sequences of state, action, next state transitions) and expresses which of the two they prefer as a solution to a given task of interest. A reward function is then learnt to explain the pattern of preferences. This approach is popular in the existing literature \citep{wirth2016model,christiano2017deep,lee2021pebble} and has a firm psychological basis. Experimental results indicate that humans find it cognitively easier to make relative (\textit{vs.} absolute) quality judgements \citep{kendall1975kendall,wilde2020improving} and exhibit lower variance when doing so \citep{guo2018experimental}. This is due in part to the lack of requirement for an absolute scale to be maintained in working memory, which is liable to induce bias as it shifts over time \citep{eric2007active}.

\begin{figure}[t]
\centering
\includegraphics[width=1\textwidth]{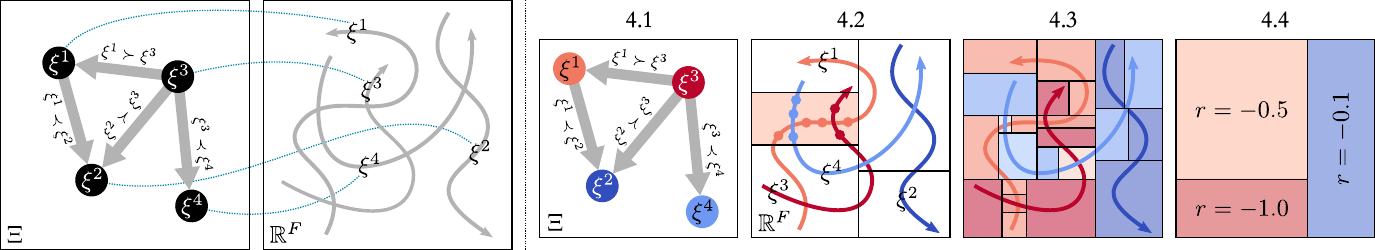}
\vspace{-0.65cm}
\caption{Left: The input to preference-based reward learning is a directed graph over a trajectory set $\Xi=\{\xi^i\}_{i=1}^N$, where each edge $(i,j)$ represents a preference $\xi^j\succ\xi^i$.
Each member of $\Xi$ is a sequence in $\mathbb{R}^F$ (blue connectors show mapping). Right: The four model induction stages; return estimation (Section \ref{sec:traj_level}), leaf-level reward prediction (\ref{sec:reg_dataset}), tree growth (\ref{sec:growth}), and pruning (\ref{sec:pruning}).}
\vspace{-0.45cm}
\label{fig:problem_and_solution}
\end{figure}

\vspace{-0.1cm}
We formalise a trajectory $\xi^i$ as a sequence $(\textbf{x}^i_1\sdot\textbf{x}^i_{T^i})$, where $\textbf{x}^i_t=\phi(s^i_{t-1},a^i_{t-1},s^i_t)\in\mathbb{R}^F$ represents a single transition as an $F$-dimensional feature vector. Given $N$ trajectories, $\Xi=\{\xi^i\}_{i=1}^N$, we assume the human provides $K\leq N(N-1)/2$ pairwise preference labels, $\mathcal{L}=\{(i,j)\}_{k=1}^K$, each of which indicates that the $j$th trajectory is preferred to the $i$th (denoted by $\xi^j\succ\xi^i$). Figure \ref{fig:problem_and_solution} (left) shows how a preference dataset $\mathcal{D}=(\Xi,\mathcal{L})$ can be viewed as a directed graph.

To learn a reward function from $\mathcal{D}$, we must assume a generative model for the preference labels. Typically, it is assumed that the human produces labels in Boltzmann-rational accordance with the sum of rewards (or \textit{return}) output by a latent reward function over the feature space, $R:\mathbb{R}^F\rightarrow\mathbb{R}$. This is formalised by adapting the classic preference model of \citet{bradley1952rank}:
\vspace{-0.175cm}
\begin{equation}
    \label{eq:bradley-terry}
    P(\xi^j\succ\xi^i|R)=
    \frac{1}{1+\exp(\frac{1}{\beta}(G(\xi^i|R)-G(\xi^j|R)))},
    \ \ \ \text{where}\ \ \ G(\xi^i|R)=\sum\nolimits_{t=1}^{T^i}R(\textbf{x}^i_t),
    \vspace{-0.125cm}
\end{equation}
and $\beta>0$ is a
temperature
coefficient. The objective of reward learning is to approximate $R$ within some learnable function class $\mathcal{R}$. This is often formalised as minimising the negative log-likelihood (NLL) loss over $\mathcal{L}$.
\citet{wirth2016model} also use the discrete $0$-$1$ loss, which considers only the directions of predicted preferences rather than their strengths. These two losses are defined as:
\vspace{-0.15cm}
\begin{equation}
    \label{eq:loss_functions}
    \ell_{\text{NLL}}(\mathcal{D},R)=
    \sum_{(i,j)\in\mathcal{L}}-\log P(\xi^j\succ\xi^i|R); \ \ \ \ \ 
    \ell_{\text{0-1}}(\mathcal{D},R)=
    \sum_{(i,j)\in\mathcal{L}}\mathbb{I}[
    P(\xi^j\succ \xi^i|R) \leq 0.5
    ].
\end{equation}

\vspace{-0.5cm}
\section{Reward Tree Induction}
\label{sec:model_induction}

\vspace{-0.2cm}
In prior work, the function class $\mathcal{R}$ has been that of linear models $R(\textbf{x})=\textbf{w}\cdot\textbf{x}$ \citep{sadigh2017active}, which have very limited expressivity, or deep NNs \citep{christiano2017deep}, which resist human interpretation. As an intermediate option, \citet{bewley2022interpretable} (henceforth BL) propose the reward tree model. Here, the parameter space consists of node-level splitting rules and reward predictions for an axis-aligned decision tree, whose leaves induce a hyperrectangular partition of $\mathbb{R}^F$.
These parameters are non-differentiable, making end-to-end optimisation of the losses in Equation \ref{eq:loss_functions} computationally intractable. BL introduce a bespoke, multi-stage model induction method that greedily optimises a proxy objective at each stage. The four stages outlined in this section, and depicted in Figure \ref{fig:problem_and_solution} (right), represent a refinement of this approach with simplified notation.

\vspace{-0.3cm}
\subsection{Trajectory-level Return Estimation}
\label{sec:traj_level}

\vspace{-0.2cm}
The stage considers the $N$ trajectories as atomic units, and uses the preference graph to construct a vector of return estimates $\textbf{g}\in\mathbb{R}^N$, which should be higher for more commonly preferred trajectories (blue in Figure \ref{fig:problem_and_solution} (4.1), c.f. red). This is a vanilla preference learning problem of the kind routinely faced in AI, psychology and economics, and thus admits a standard solution. BL's original proposal finds the least squares solution for $\textbf{g}$ under Thurstone's Case V preference model \citep{gulliksen1956least}. For consistency with prior work, we instead minimise the NLL loss under the Bradley-Terry model (we find it makes little difference to the result). Concretely, the proxy objective for this stage is
\vspace{-0.15cm}
\begin{equation}
    \label{eq:proxy_loss_trajectory_level}
    \underset{\textbf{g}\in\mathbb{R}^N}{\text{argmin}}\Big[\sum_{(i,j)\in\mathcal{L}}-\log \frac{1}{1+\exp(\textbf{g}^i-\textbf{g}^j)}\Big],\ \ \ \text{subject to}\ \ \ \left\{\begin{array}{r}\text{min}(\textbf{g})=0\\\text{or }\text{max}(\textbf{g})=0\end{array}\right.\ \text{and}\ \text{std}(\textbf{g})=\beta,
    \vspace{-0.2cm}
\end{equation}
where $\beta$ is the mean trajectory length in $\Xi$, $\sum_{i=1}^N T^i/N$. The min-or-max constraint ensures that all return estimates have the same sign (positive or negative), which aids both policy learning and interpretability (see Appendix \ref{app:sign_constraint}). We optimise Equation \ref{eq:proxy_loss_trajectory_level} by unconstrained gradient descent with the Adam optimiser \citep{kingma2014adam}, followed by post-normalisation to meet the constraints.

\vspace{-0.3cm}
\subsection{Leaf-level Reward Estimation}
\label{sec:reg_dataset}

\vspace{-0.2cm}
The vector $\textbf{g}$ estimates trajectory-level returns, but the aim of reward learning is to decompose these into sums of rewards for the constituent transitions. BL's novel contribution is to do this using a tree model $\mathcal{T}$, consisting of a hierarchy of rules that partition the transition-level feature space $\mathbb{R}^F$ into $L_\mathcal{T}$ hyperrectangular subsets called \textit{leaves}. Let $\text{leaf}_\mathcal{T}:\mathbb{R}^F\rightarrow\{1..L_\mathcal{T}\}$ be a function that maps a feature vector $\textbf{x}\in\mathbb{R}^F$ to the leaf in which it resides by propagating it through the rule hierarchy.
The predicted reward for any $\textbf{x}$ lying in leaf $l$ is defined as an average over $\textbf{g}$, weighted by the proportion of timesteps that each trajectory in $\Xi$ spends in $l$:
\vspace{-0.2cm}
\begin{equation}
    \label{eq:reward_prediction}
    R_\mathcal{T}(\textbf{x})
    =\textbf{r}_{\text{leaf}(\textbf{x})},
    \ \ \ \text{where}\ \ \ 
    \textbf{r}_l=\sum_{i=1}^N\frac{\textbf{g}^i}{T^i}\frac{\sum_{t=1}^{T^i}\mathbb{I}[\text{leaf}_\mathcal{T}(\textbf{x}^i_t)=l]}{\sum_{i=1}^N\sum_{t=1}^{T^i}\mathbb{I}[\text{leaf}_\mathcal{T}(\textbf{x}^i_t)=l]},\ \forall\ 1\leq l\leq L_\mathcal{T}. 
    \vspace{-0.1cm}
\end{equation}
In essence, the effect of Equation \ref{eq:reward_prediction} is to predict higher reward in leaves that contain more timesteps from trajectories with high $\textbf{g}$ values. While ostensibly na{\"i}ve, BL find that this time-weighted credit assignment is more robust than several more sophisticated alternatives. It reduces the number of free parameters in subsequent induction stages, permits fast implementation, and provides an intuitive interpretation of predicted reward that is traceable back to the contribution of each $\xi^i\in\Xi$. Figure \ref{fig:problem_and_solution} (4.2) shows how timesteps from $\xi^1$, $\xi^3$ and $\xi^4$ contribute to the reward prediction for one leaf.

\vspace{-0.3cm}
\subsection{Tree Growth}
\label{sec:growth}

\vspace{-0.2cm}
Recall that the objective of preference-based reward learning is to find a reward model that optimises a measure of fidelity to $\mathcal{D}$, such as the losses in Equation \ref{eq:loss_functions}. When the model is a tree, this is achieved by the discrete operations of growth (adding partitioning rules) and pruning (removing rules). Given a tree $\mathcal{T}$, a new rule has the effect of splitting the $l$th leaf with a hyperplane at a location $c\in\mathcal{C}_f$ along the $f$th feature dimension (where $\mathcal{C}_f\subset\mathbb{R}$ is a set of candidate split thresholds, e.g. all midpoints between unique values in $\Xi$). Let $\mathcal{T}+[lfc]$ denote the newly-enlarged tree. Splitting recursively creates an increasingly fine partition of $\mathbb{R}^F$. Figure \ref{fig:problem_and_solution} (4.3) shows an example with 23 leaves.

\vspace{-0.15cm}
A central issue is the criterion for selecting the next rule to add. BL use the proxy objective of minimising the local variance of reward predictions, which is exactly the CART algorithm \citep{breiman2017classification}. While very fast, this criterion is only loosely aligned with fidelity to $\mathcal{D}$. We additionally consider the more direct criterion of greedily maximising the immediate reduction in $\ell_\text{0-1}$:
\vspace{-0.1cm}
\begin{equation}
    \label{eq:split_criterion}
    \text{argmax}_{1\leq l\leq L_\mathcal{T},\ 1\leq f\leq F,\ c\in\mathcal{C}_f}\ \left[\ \ell_{\text{0-1}}(\mathcal{D},R_\mathcal{T})-\ell_{\text{0-1}}(\mathcal{D},R_{\mathcal{T}+[lfc]})\ \right].
    \vspace{-0.1cm}
\end{equation}
In Section \ref{sec:fid_perf_results}, we show that switching to this criterion consistently improves performance on most quantitative measures (we also tried a criterion based on $\ell_\text{NLL}$, but found it to be far more computationally costly and prone to overfitting). Recursive splitting stops when either no reduction in $\ell_\text{0-1}$ can be achieved by any single split, or a tree size limit $L_\mathcal{T}=L_\text{max}$ is reached.

\vspace{-0.3cm}
\subsection{Tree Pruning}
\label{sec:pruning}

\vspace{-0.2cm}
Growth is followed by a pruning sweep which reduces the size of the tree by rule removal. Such reduction is beneficial for both performance (\citet{tien2022study} find that increased model capacity raises the risk of causal confusion in preference-based reward learning) and human comprehension (in the language of \citet{jenner2022preprocessing}, it is a form of \textit{``processing for interpretability"}). Given a tree $\mathcal{T}$, one pruning operation has the effect of merging two leaves into one by removing the rule at the common parent node. Let $\mathbb{T}$ denote the sequence of nested subtrees induced by pruning the tree recursively back to its root, at each step removing the rule that minimises the next subtree's $\ell_{\text{0-1}}$.
We
select the $\mathcal{T}\in\mathbb{T}$ that minimises $\ell_{\text{0-1}}$, additionally regularised by a term that encourages small trees: $\text{argmin}_{\mathcal{T}\in\mathbb{T}}[\ell_{\text{0-1}}(\mathcal{D},R_\mathcal{T})+\alpha L_\mathcal{T}]$, where $\alpha\geq 0$. Note that even with $\alpha=0$ pruning may still yield a reduced tree,
as unlike in traditional decision tree induction, the effect of individual rules on $\ell_{\text{0-1}}$ depends on the order in which they are added or removed. In the example in Figure \ref{fig:problem_and_solution} (4.4), pruning yields a final tree with $3$ leaves, for which indicative leaf-level reward predictions are shown.

\vspace{-0.3cm}
\section{
Online Learning Setup
}
\label{sec:online_learning}

\vspace{-0.2cm}
\subsection{Iterated Policy and Reward Learning}

\vspace{-0.2cm}
Sections \ref{sec:pbrl} and \ref{sec:model_induction} do not discuss the origins of the trajectories $\Xi$, or how reward learning relates to the downstream objective of learning a policy for the underlying task. Following most recent work since \citet{christiano2017deep}, we resolve both questions with an online bootstrapped approach. Assuming an episodic MDP, the $i$th episode of policy learning produces a new trajectory $\xi^i$ to add to $\Xi$. We immediately connect $\xi^i$ to the preference graph by asking the human to compare it to $K_\text{batch}$ random trajectories from the existing set (while \citet{sadigh2017active} and others have proposed active querying schemes, that is not our focus here,
and this simple strategy performs satisfactorily). We then update the reward tree on the full preference graph via the four stages given in Section \ref{sec:model_induction}. We find that BL's original method of starting growth from the current state of the tree causes lock-in to poor initial solutions,
so instead re-grow from scratch on each update. The rule structure nonetheless tends to stabilise, as the enlarging preference graph becomes increasingly similar for later updates.
For the $(i+1)$th episode, the policy learning agent then attempts to optimise for the newly-updated reward. 
By iterating this process up to a total preference budget $K_\text{max}$ and/or episode budget $N_\text{max}$,
we hope to converge to both a reward
tree that reflects the human's preferences, and an agent policy that satisfies those preferences. Appendix \ref{app:pseudocode} contains pseudocode for the online algorithm.

\vspace{-0.3cm}
\subsection{Integration with Model-based RL}
\label{sec:model_based}

\vspace{-0.2cm}
Reward learning methods are generally agnostic to the structure of the policy learning agent; this modularity is hailed as an advantage over other human-agent teaching paradigms \citep{leike2018scalable}. In line with most recent works, BL use a model-free RL agent, specifically soft actor-critic (SAC) \citep{haarnoja2018soft}. However, other works \citep{reddy2020learning,rahtz2022safe} use model-based RL (MBRL) agents that leverage learnt dynamics models and planning. MBRL is attractive in the reward learning context because it disentangles the predictive and normative aspects of decision-making. Since (assuming no changes to the environment) dynamics remain stationary during online reward learning, the amount of re-learning required is reduced and along with it, the risk of pitfalls such as manipulation \citep{armstrong2020pitfalls} and premature convergence. MBRL agents also provide a richer target for XAI methods due to \textit{``the explicit nature of their world knowledge and of the reasoning performed to take decisions"} \citep{hoffmann2019explainable}. In future work, we plan to explore how this could
be synergistic with the adoption of reward trees (see Section \ref{sec:conclusion}). Finally, MBRL can be very data-efficient; we find that switching from SAC to a model-based algorithm called PETS \citep{chua2018deep} reduces training steps by multiple orders of magnitude, and cuts wall-clock runtime (see Appendix \ref{app:sac}). PETS selects actions by decision-time planning through a learnt dynamics model $D':\mathcal{S}\times\mathcal{A}\rightarrow\Delta(\mathcal{S})$ up to a horizon $H$. In state $s$, planning searches for a sequence of $H$ future actions that maximise return under the current reward model:
\vspace{-0.03cm}
\begin{equation}
    \label{eq:pets_action_sequence_eval}
    \underset{(a_0,...,a_{H-1})\in\mathcal{A}^H}{\text{argmax}}\mathbb{E}_{D'}\Big[\sum\nolimits_{h=0}^{H-1}\gamma^h R_\mathcal{T}(\phi(s_h,a_h,s_{h+1}))\Big],\ \text{where}\ s_0=s,\ s_{h+1}\sim D'(s_h,a_h).\
    \vspace{-0.03cm}
\end{equation}
The first action $a=a_0$ is executed, and then the agent re-plans on the next timestep.
In practice, $D'$ is an ensemble of probabilistic NNs, the expectation over $D'$ is replaced by a Monte Carlo estimate, and the optimisation is approximated by the iterative cross-entropy method.

\vspace{-0.3cm}
\section{Experiments and Results}
\label{sec:results}

\vspace{-0.2cm}
In this section, we combine quantitative and qualitative evaluations to assess the performance of reward tree learning, specifically in comparison to the standard approach of using NNs. We also illustrate how the intrinsic interpretability of reward trees allows us to analyse what they have learnt.

\begin{figure}[h!]
\centering
\includegraphics[width=1\textwidth]{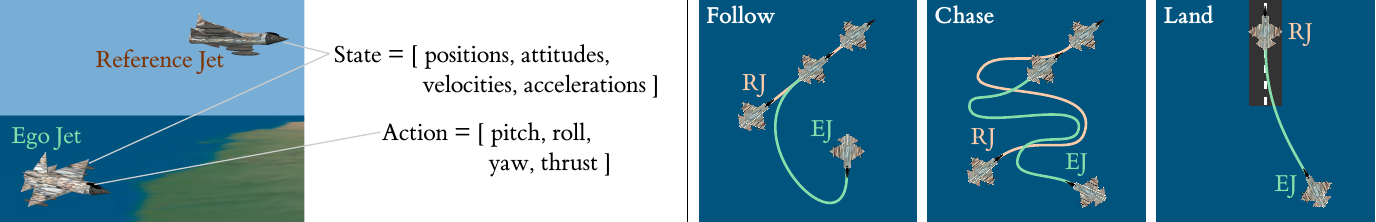}
\vspace{-0.65cm}
\caption{Handling tasks.
Follow: turn to fly in formation with RJ on a linear path. Chase: maintain distance/line of sight to RJ as it turns randomly. Land: approach a runway using RJ as a reference.}
\vspace{-0.15cm}
\label{fig:fastjet}
\end{figure}

\vspace{-0.15cm}
Our experiments consider three episodic tasks in an aircraft handling environment, in which the agent must manoeuvre an aircraft (the \textit{ego jet}, EJ) in a desired manner relative to a second \textit{reference jet} (RJ) whose motion, if any, is considered part of the environment dynamics. These tasks provide useful test cases for reward learning because each has a large space of plausible reward functions, which may reflect the divergent priorities
and stylistic preferences of domain experts.
Such expert knowledge is often tacit and difficult to codify by hand \citep{sternberg1999tacit}, motivating a data-driven approach.
Figure \ref{fig:fastjet} shows a schematic of each task, and Appendix \ref{app:env_details} contains a broader justification for this experimental domain alongside task and implementation details.


\vspace{-0.14cm}

In place of costly human-in-the-loop evaluation, our experiments use synthetic \textit{oracle} preferences with respect to nominal reward functions of varying complexity, which are given in Appendix \ref{app:tasks_and_oracles}. This approach is popular \citep{griffith2013policy,christiano2017deep,reddy2020learning,lindner2021information} as it enables scalable systematic comparison, with the ability to quantify performance (and in our case, appraise
learnt trees) in terms of reconstruction of a known ground truth. However, emulating a human with an oracle that responds with perfect rationality is unrealistic \citep{lee2021bpref}. For this reason, Section \ref{sec:sensitivity_analysis} examines the performance impacts of noisy and myopic oracles, and a restricted data budget. Experimental details and hyperparameters are given in Appendix \ref{app:exp_details}.



\subsection{Quantitative Performance}
\label{sec:fid_perf_results}
\vspace{-0.2cm}

We evaluate online reward learning with PETS using trees with both $\ell_{\text{0-1}}$-based and variance-based split criteria, baselined against the de facto standard of using NNs
(see Appendix \ref{app:nn_baseline} for details). We use $K_\text{max}=1000$ preferences over $N_\text{max}=200$ online trajectories, and perform $10$ repeats.
As a headline statistic, we compute the \textit{oracle regret ratio} (ORR), which is the drop in median oracle return of PETS agents deployed using each trained reward model compared with direct access to the oracle reward, as a fraction of the drop to a random policy (lower is better). Below are the median (top) and minimum (bottom) ORR values across the $10$ repeats for each task-model combination:

\vspace{-0.325cm}
\begin{table}[h!]
\centering
\small
\begin{tabular}{c|c|c|c|c|c|c|c|c}
\multicolumn{3}{c|}{Follow}    & \multicolumn{3}{c|}{Chase}    & \multicolumn{3}{c}{Land}     \\
NN & Tree(0-1) & Tree (var) & NN & Tree (0-1) & Tree (var) & NN & Tree (0-1) & Tree (var) \\
\hline
$.000$ & $.120$ & $.284$ & $-.030$ & $.040$ & $.126$ & $.014$ & $.050$ & $.062$ \\
$-.010$ & $.057$ & $.158$ & $-.051$ & $-.011$ & $.065$ & $-.030$ & $.011$ & $.010$
\end{tabular}
\end{table}
\vspace{-0.35cm}

We observe that:
1) NN reward learning is strong on all tasks;
2) switching to a reward tree induces a variable performance hit;
3) $\ell_{\text{0-1}}$ splitting consistently outperforms the variance-based method; and 4) both NN and tree models sometimes exceed the direct use of the oracle (negative ORR). This has been observed before \citep{cao2021weak} and may be due to improved shaping in the learnt reward.

\begin{figure}[H]
\centering
\includegraphics[width=1\textwidth]{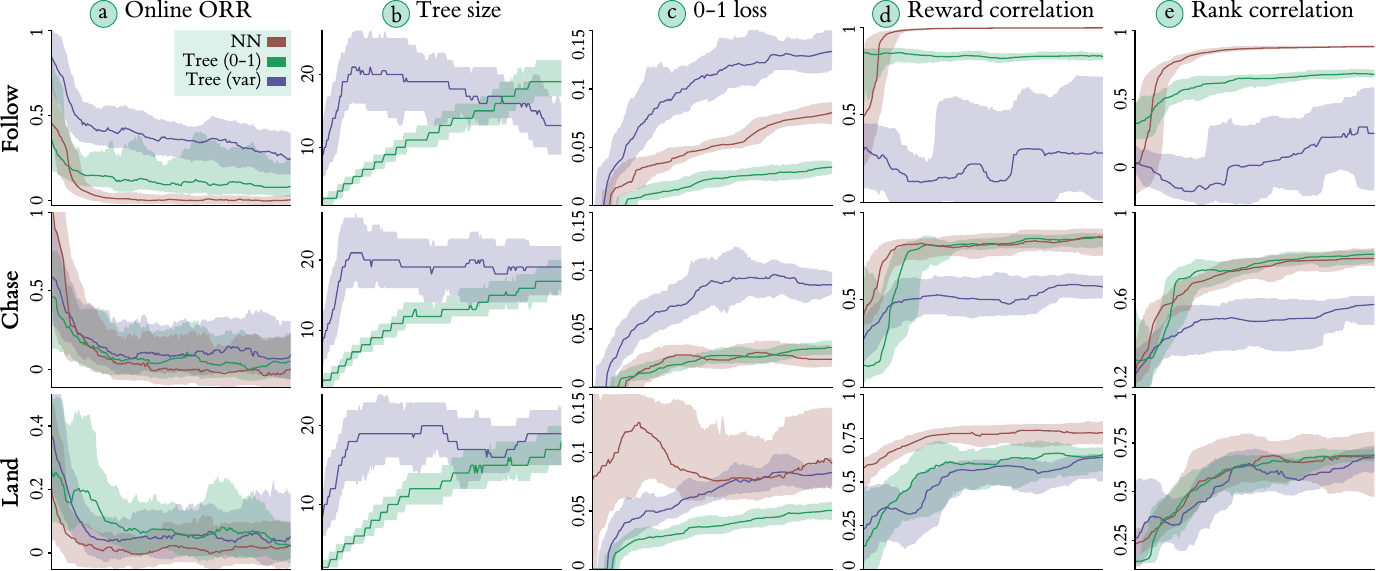}
\vspace{-0.65cm}
\caption{Time series of metrics for online NN- and tree-based reward learning on all three tasks.}
\vspace{-0.45cm}
\label{fig:results_main}
\end{figure}

\vspace{-0.15cm}
Figure \ref{fig:results_main} expands these results with more metrics, revealing subtler trends not captured by headline ORR values. Metrics are plotted as time series over the $200$ learning episodes (sliding-window medians and interquartile ranges across repeats). In the left column (\textbf{a}), the ORR of online trajectories shows how agent performance converges. For Follow, there is a gap between the models, with $\ell_{\text{0-1}}$ splitting clearly aiding performance but still lagging behind the NNs. The learning rates for Chase and Land are more homogeneous, but the NNs reach somewhat lower asymptotes. For the reward tree models, (\textbf{b}) shows how the number of leaves changes over time. The variance-based trees tend to grow rapidly initially before stabilising or shrinking, while the $\ell_{\text{0-1}}$ trees enlarge more conservatively, suggesting this method is less liable to overfit to small preference graphs. Trees of a readily-interpretable size ($\approx 20$ leaves) are produced for all tasks; it is possible that performance could be improved by independently tuning the size regulariser $\alpha$ per task. (\textbf{c}) shows $\ell_{\text{0-1}}$ over time, which tends to increase as the growing preference graph presents a harder reconstruction problem, though the shape of all curves suggests convergence (note that random prediction gives $\ell_{\text{0-1}}=0.5$). For Follow and Land, the trees that directly split on $\ell_{\text{0-1}}$ actually perform better than the NNs; they more accurately predict the direction of preferences in the graph. The fact that this does not translate into lower ORR indicates that the problems of learning a good policy and exactly replicating the preference dataset are not identical, a point made by \citet{lindner2021information}. In the final two columns, we follow \citet{gleave2021quantifying} in performing an unbiased, \textit{policy-invariant} comparison of the models by correlating their outputs with the oracle reward functions on common evaluation datasets (see Appendix \ref{app:coverage_datasets} for dataset creation). We compute online correlations with the oracles in terms of both transition-level rewards (\textbf{d}) and the ordinal ranking of trajectories by return (\textbf{e}), the latter via the \citet{kendall1938new} $\tau$ coefficient. The curves subtly differ, indicating that it is possible to reconstruct trajectory rankings (and by extension, any pairwise preferences) to a given accuracy with varying fidelity at the individual reward level. However, the common overall trend is that $\ell_{\text{0-1}}$-based trees outperform variance-based ones, with NNs sometimes improving again by a smaller margin, and sometimes bringing no additional benefit. Moving top-to-bottom down the tasks, the gap between models reduces from both sides; NN performance worsens while variance-based trees improve.




\vspace{-0.25cm}

\newpage
A potentially important factor in these experiments is that the oracle reward for Follow is a simple linear function,
while the other two contain progressively more terms and discontinuities (see Appendix \ref{app:tasks_and_oracles}). A trend suggested by these results is thus that the performance gap between NN and tree architectures (on both ORR and correlation metrics) reduces as the ground truth reward becomes more complex and nonlinear. Further experiments would be needed to test this hypothesis.

\vspace{-0.3cm}
\subsection{Visual Trajectory Inspection}

\begin{figure}[H]
\centering
\includegraphics[width=1\textwidth]{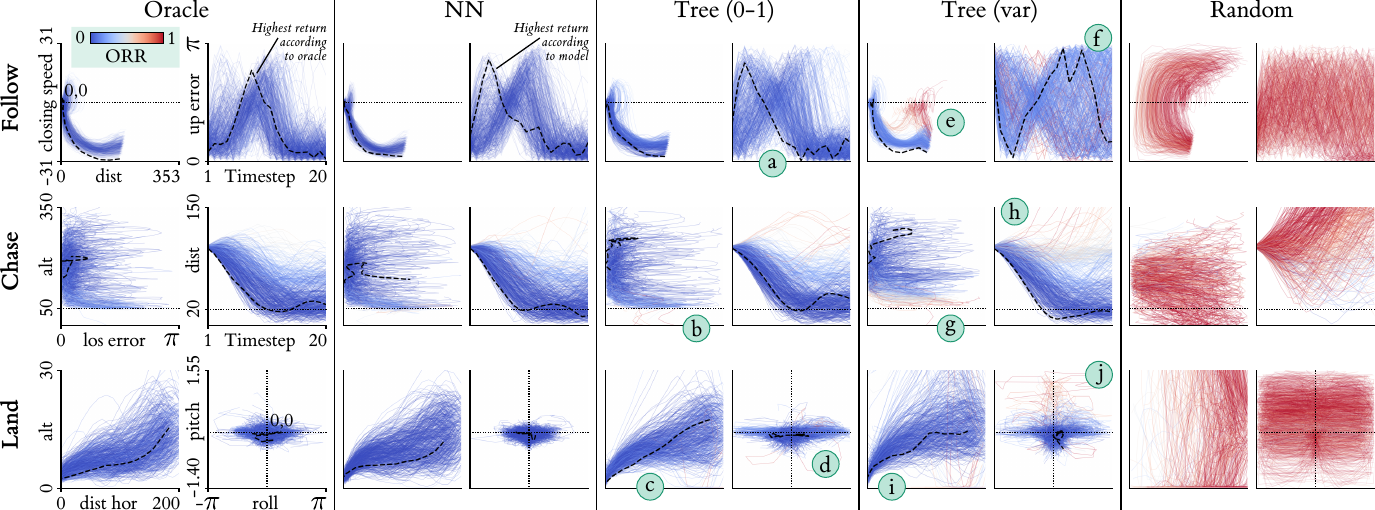}
\vspace{-0.7cm}
\caption{Agent trajectories using the best models by ORR, with oracle and random for comparison.}
\vspace{-0.45cm}
\label{fig:trajectory_visualisation}
\end{figure}

\vspace{-0.2cm}
While useful for benchmarking, quantitative metrics provide little insight into the structure of the learnt solutions. They would also mostly be undefined when learning from humans since the ground truth reward is unknown. We therefore complement them with a visual analysis of induced agent behaviour. Figure \ref{fig:trajectory_visualisation} plots $500$ trajectories of PETS agents using the best repeat by ORR for each task-model combination, across a range of 
features as well as time
(see Appendix \ref{app:env_implementation} for feature definitions). Dashed curves show the trajectory with the highest predicted return according to each model. We also show trajectories for PETS agents with direct oracle access, and for random policies.

\vspace{-0.15cm}
The high-level trend is that all models are far closer to the oracle than random, with few examples of obviously incorrect behaviour (highlighted in red, due to colouring by ORR). The NNs induce trajectories that are almost indistinguishable from the oracle, although the $\ell_{\text{0-1}}$-based reward trees lag not far behind, with the variance-based trees producing more anomalies. Successes of the $\ell_{\text{0-1}}$ trees include the execution of Follow with a single banked turn before straightening up, as shown by the \texttt{up} \texttt{error} time series (\textbf{a}).
Indeed, the trajectories for this model are almost imperceptibly different from those of the NN, a result which is belied by the mediocre ORR of $0.158$. This underlines the importance of joint quantitative-qualitative evaluation. For Chase (\textbf{b}), the $\ell_{\text{0-1}}$ tree has learnt to keep the agent narrowly above the altitude threshold $\texttt{alt}<50$, below which the oracle reward is strongly negative (see Appendix \ref{app:tasks_and_oracles}). The threshold is violated in only eight of $500$ trajectories ($1.6\%$). For Land, the $\ell_{\text{0-1}}$ tree replicates the oracle in producing a gradual reduction in \texttt{alt} (\textbf{c}) while usually keeping \texttt{pitch} close to $0$ (\textbf{d}), although the distribution of \texttt{roll} values is less narrow.

\vspace{-0.15cm}
In contrast, the agent using the variance-based tree for Follow sometimes fails to reach the target position (\textbf{e}; red trajectories), and also does not reliably straighten up to reduce \texttt{up} \texttt{error} (\textbf{f}). For Chase, the altitude threshold does not appear to have been learnt precisely, and lower-altitude trajectories often fail to close the distance to RJ (\textbf{g} and \textbf{h}; red trajectories). For Land, the variance-based tree gives a later and less smooth descent (\textbf{i}), and less consistent pitch control (\textbf{j}), than the NN or $\ell_{\text{0-1}}$-based tree, although all models produce a somewhat higher altitude profile than the oracle.

\vspace{-0.25cm}

\subsection{Sensitivity Analysis}
\label{sec:sensitivity_analysis}

\begin{figure}
\centering
\includegraphics[width=1\textwidth]{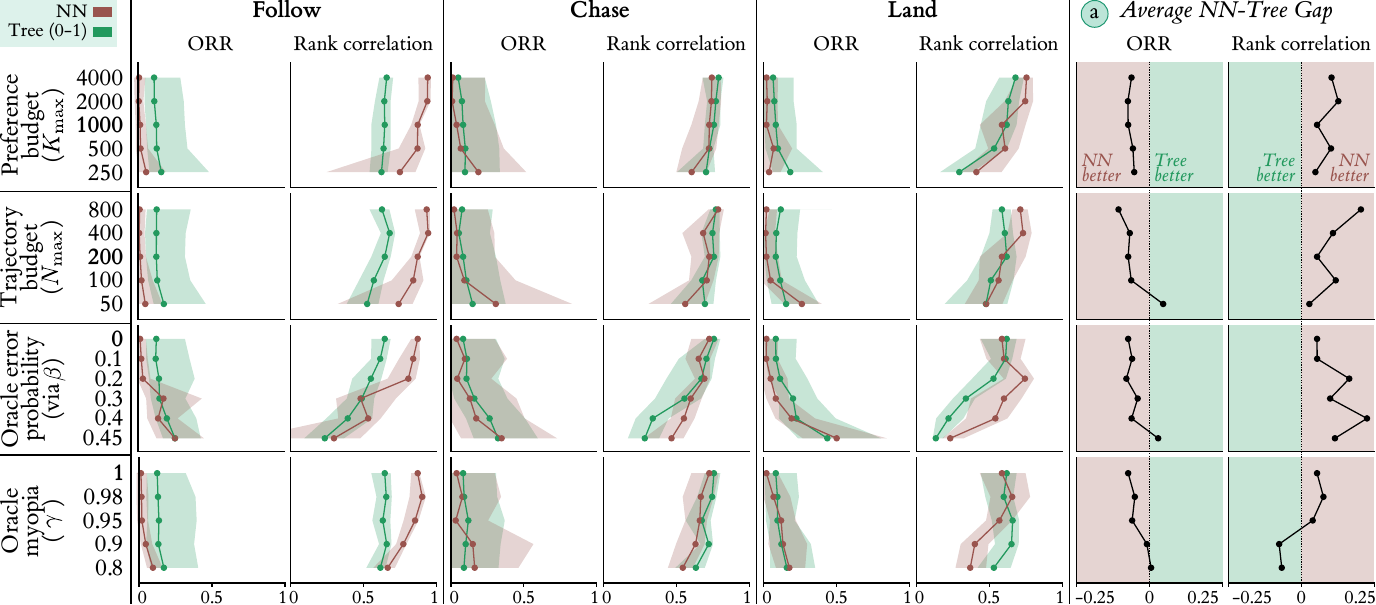}
\vspace{-0.65cm}
\caption{Comparative sensitivity analysis of reward learning with NNs and trees.}
\vspace{-0.55cm}
\label{fig:sensitivity_analysis}
\end{figure}

\vspace{-0.2cm}
It is important to consider how learning performance degrades with reduced or corrupted data. In Figure \ref{fig:sensitivity_analysis}, we evaluate the effect of varying the number of preferences $K_\text{max}$ (with fixed $N_\text{max}=200$) and trajectories $N_\text{max}$ (with fixed $K_\text{max}=1000$) on reward learning with NNs and $\ell_{\text{0-1}}$-splitting trees. Following \citet{lee2021bpref}, we also create more human-like preference data via two modes of oracle \textit{irrationality}: preference noise (by using a nonzero Boltzmann temperature $\beta$ to give a desired error rate on the coverage datasets) and a myopic recency bias (by exponentially discounting earlier timesteps when evaluating trajectory returns). We run five repeats for all cases, and report the medians and interquartile ranges of ORR (lower is better) and rank correlation (higher is better).

\vspace{-0.15cm}
Both NN and tree models exhibit good robustness with respect to all four parameters. Although NNs remain superior in most cases, the gap varies, and is often reduced compared to the base cases (bold labels). The budget sensitivity is low, with little improvement for $K_\text{max}>1000$ and $N_\text{max}>200$, and no major drop even with $25\%$ of the data as the base case. For all tasks, the oracle error probability can increase to around $20\%$ before significant drops in performance are observed. This is a promising indicator of the transferability of reward tree learning to imperfect human data. Another general observation is that the trends for trees are somewhat smoother than for NNs, with fewer sharp jumps and fewer instances of very high spread across the five repeats.

\vspace{-0.15cm}
In the right column (\textbf{a}), we aggregate these results by taking the difference between the NN and tree metrics, and averaging across the three tasks. In all cases aside from rank correlation with $\beta>0$, the NN-tree performance gap tends to become more favourable to the tree models as the varied parameter becomes more challenging (top-to-bottom). This sensitivity analysis thus indicates that trees are at least as robust to difficult learning scenarios as NNs, and may even be slightly more so.

\vspace{-0.2cm}

\vspace{-0.2cm}
\subsection{Tree Structure Analysis}
\label{sec:tree_analysis}

\begin{figure}[t]
\centering
\includegraphics[width=1\textwidth]{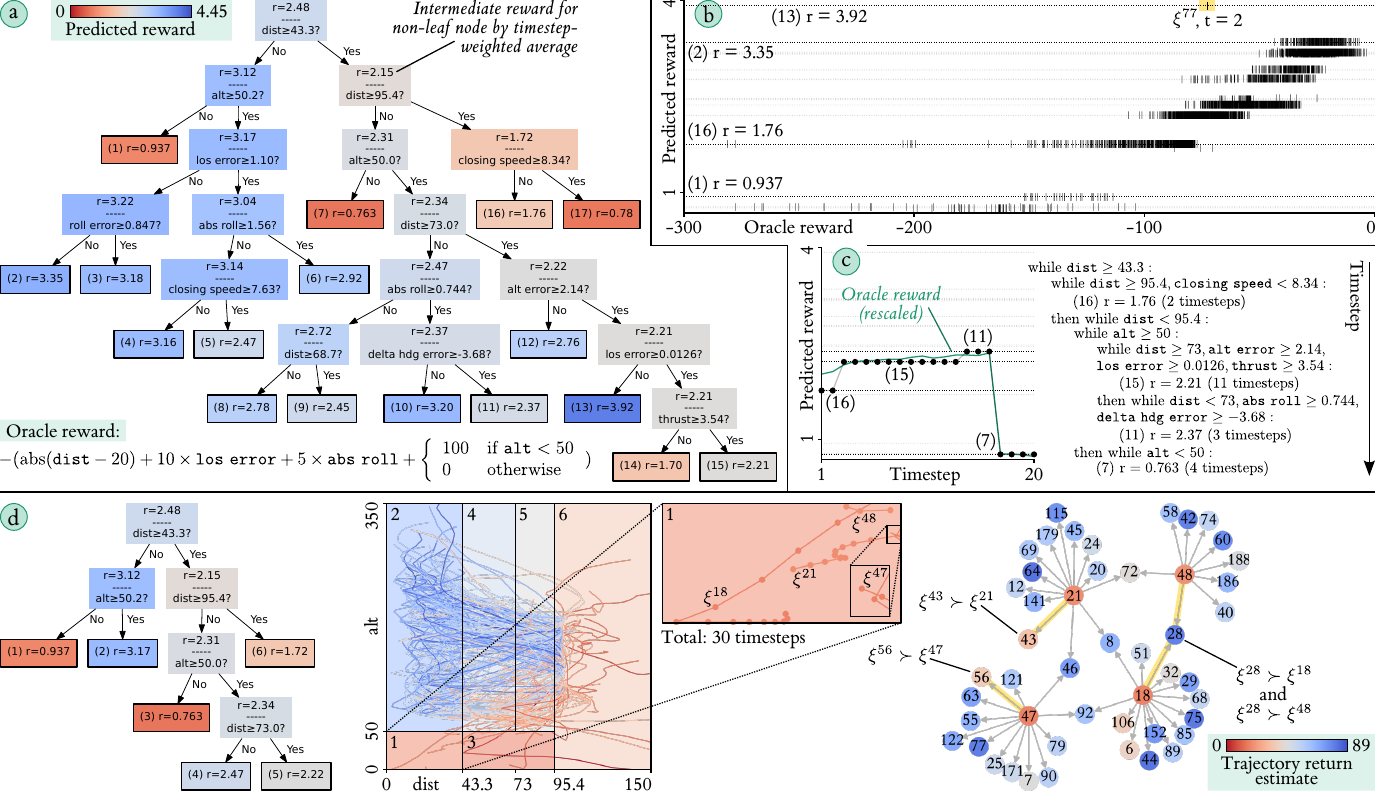}
\vspace{-0.65cm}
\caption{Analysis of a reward tree learnt for the Chase task.}
\vspace{-0.5cm}
\label{fig:tree_analysis}
\end{figure}

\vspace{-0.2cm}
Thus far we have shown that reward learning with $\ell_{\text{0-1}}$-based trees can be competitive with NNs, but not quite as performant overall. We now turn to a concrete advantage which may tip practical
tradeoffs
in its favour: the ability to interpret the learnt model, and analyse how its structure arises from the underlying preference graph. In this section we favour depth over breadth, so focus on the single best tree by ORR on the Chase task. The analysis in Figure \ref{fig:tree_analysis} is divided into sections (\textbf{a} -- \textbf{d}):

\vspace{-0.15cm}
(\textbf{a})\quad This reward tree has $17$ leaves.
The oracle reward, printed below, uses four features, all of which are used in the tree in ways that are broadly aligned (e.g. lower \texttt{los} \texttt{error} leads to leaves with higher reward). The model has learnt the crucial threshold $\texttt{alt}<50$, correctly assigning low reward when it is crossed. This explains why we observe rare violations of the altitude threshold in Figure \ref{fig:trajectory_visualisation}. However, it has not learnt the ideal distance to RJ, $\texttt{dist}=20$, with $43.3$ being the lowest value used in a rule. This could be because the underlying preference graph lacks sufficient preferences to make this distinction; adopting an active querying scheme may help to discover such subtleties efficiently. Other features besides those used by the oracle are present in the tree, indicating some causal confusion \citep{tien2022study}. This may not necessarily harm agent performance, as it could provide beneficial shaping (e.g. penalising positive \texttt{closing} \texttt{speed}, which indicates increasing distance to RJ). That may indeed be the case for this model since ORR is actually negative.

\vspace{-0.15cm}
(\textbf{b})\quad We plot the tree's predicted reward against the oracle reward for all timesteps in the online trajectories (correlation $=0.903$). The predictions for each leaf lie along a horizontal line. Most leaves, including $1$ and $2$, are well-aligned on this data because their oracle reward distributions are tightly concentrated around low/high averages respectively (note that the absolute scale is irrelevant here). Leaf $16$ has a wider oracle reward distribution, with several negative outliers. An optimal tree would likely split this leaf further, perhaps using the $\texttt{alt}<50$ threshold.
The one anomaly is leaf $13$, which contains just a single timestep from $\xi^{77}$. This trajectory is the eighth best in the dataset by oracle return, but this leaf assigns that credit to a state that seemingly does not merit it, as the distance to RJ is so high ($\texttt{dist}>73$). This may be an example of suboptimal tree induction, but the fact that its origin can be pinpointed precisely is a testament to the value of interpretability.

\vspace{-0.15cm}
(\textbf{c})\quad We leverage the tree structure to produce a human-readable explanation of reward predictions for a single trajectory, which may be of value to an end user (e.g. a pilot). We consider $\xi^{191}$, a rare case that violates the altitude threshold. The time series of reward shows that the $20$ timesteps are spent in leaves $16$, $15$, $11$ and $7$. Rescaled oracle rewards are overlaid in teal, and show that the model's predictions are well-aligned. To the right, we translate this visualisation into a textual form, similar to a nested program. Read top-to-bottom, the text indicates which rules of the tree are active at each timestep, and the effect this has on predicted reward. This trajectory starts fairly positively, with reward gradually increasing over the first $16$ timesteps as \texttt{dist} is reduced to between $43.3$ and $73$, but then falls dramatically when the $\texttt{alt}<50$ threshold is crossed. We are unaware of any method that could extract such a compact explanation of sequential predictions from an NN.

\vspace{-0.125cm}
(\textbf{d})\quad We isolate a subtree, starting at the root node, that splits only on \texttt{dist} and \texttt{alt}. We give a spatial representation of the subtree, and how it is populated by the $200$ online trajectories, using a 2D partition plot analogous to those in Figure \ref{fig:problem_and_solution}. Zooming into leaf $1$, which covers cases where the altitude threshold is violated, we see that it contains a total of $30$ timesteps across four trajectories. By Equation \ref{eq:reward_prediction}, the low reward for this leaf results from a weighted average of the return estimates for these four trajectories, which in turn (by Equation \ref{eq:proxy_loss_trajectory_level}) are derived from the preference graph. We can use this inverse reasoning to ask \textit{why} this leaf has much lower reward than its sibling (leaf $2$ of the subtree). A proximal explanation comes by filtering the graph for preferences that specifically compare trajectories that visit those two leaves. $49$ such preferences exist, and in all cases, the oracle prefers the trajectory that does not visit leaf $1$. Some of these preferences may be more practically salient than others. For example, we might highlight trajectories that feature more than once (e.g. $\xi^{28}$ is preferred to both $\xi^{18}$ and $\xi^{48}$), or cases where trajectories with low overall return estimates are nonetheless preferred to those in leaf $1$ (e.g. $\xi^{43}\succ \xi^{21}$ and $\xi^{56}\succ \xi^{47}$). We believe that much more could be done to extend this framework for traceable explanation of preference-based reward.

\vspace{-0.35cm}
\section{Conclusion and Future Work}
\label{sec:conclusion}

\vspace{-0.2cm}
Reward learning with trees provides a promising alternative to black-box NNs, and could enable more trustworthy and verifiable agent alignment. Through oracle experiments on high-dimensional tasks, we show that reward trees with around $20$ leaves can achieve quantitative and qualitative performance close to that of NNs, with a more direct split criterion bringing consistent improvements. We find evidence that the NN-tree gap reduces as the ground truth reward becomes more nonlinear, and remains stable or reduces further in the presence of limited or corrupted data. While practical applications may accept some loss in performance for a gain in interpretability, further algorithmic improvements should be sought, including to move beyond locally-greedy split criteria. However, our immediate aim is to develop an end-to-end framework for explainable model-based agents with preference-based reward trees (roughly: planning can be reframed as comparing alternative paths through the discrete leaves of the tree).
Having established this framework, we then intend to evaluate reward learning and explanation with real human preferences in the aircraft handling domain.



\newpage
\bibliography{bibliography}

\begin{thebibliography}{62}
\providecommand{\natexlab}[1]{#1}
\providecommand{\url}[1]{\texttt{#1}}
\expandafter\ifx\csname urlstyle\endcsname\relax
  \providecommand{\doi}[1]{doi: #1}\else
  \providecommand{\doi}{doi: \begingroup \urlstyle{rm}\Url}\fi

\bibitem[Amir \& Amir(2018)Amir and Amir]{amir2018highlights}
Dan Amir and Ofra Amir.
\newblock Highlights: Summarizing agent behavior to people.
\newblock In \emph{Proceedings of the 17th International Conference on
  Autonomous Agents and MultiAgent Systems}, pp.\  1168--1176, 2018.

\bibitem[Armstrong et~al.(2020)Armstrong, Leike, Orseau, and
  Legg]{armstrong2020pitfalls}
Stuart Armstrong, Jan Leike, Laurent Orseau, and Shane Legg.
\newblock Pitfalls of learning a reward function online.
\newblock In Christian Bessiere (ed.), \emph{Proceedings of the Twenty-Ninth
  International Joint Conference on Artificial Intelligence, {IJCAI-20}}, pp.\
  1592--1600. International Joint Conferences on Artificial Intelligence
  Organization, 7 2020.
\newblock \doi{10.24963/ijcai.2020/221}.
\newblock URL \url{https://doi.org/10.24963/ijcai.2020/221}.
\newblock Main track.

\bibitem[Bajcsy et~al.(2017)Bajcsy, Losey, O’Malley, and
  Dragan]{bajcsy2017learning}
Andrea Bajcsy, Dylan~P Losey, Marcia~K O’Malley, and Anca~D Dragan.
\newblock Learning robot objectives from physical human interaction.
\newblock In \emph{Conference on Robot Learning}, pp.\  217--226. PMLR, 2017.

\bibitem[Bastani et~al.(2018)Bastani, Pu, and
  Solar-Lezama]{bastani2018verifiable}
Osbert Bastani, Yewen Pu, and Armando Solar-Lezama.
\newblock Verifiable reinforcement learning via policy extraction.
\newblock In S.~Bengio, H.~Wallach, H.~Larochelle, K.~Grauman, N.~Cesa-Bianchi,
  and R.~Garnett (eds.), \emph{Advances in Neural Information Processing
  Systems}, volume~31. Curran Associates, Inc., 2018.
\newblock URL
  \url{https://proceedings.neurips.cc/paper/2018/file/e6d8545daa42d5ced125a4bf747b3688-Paper.pdf}.

\bibitem[Bewley \& Lecue(2022)Bewley and Lecue]{bewley2022interpretable}
Tom Bewley and Freddy Lecue.
\newblock Interpretable preference-based reinforcement learning with
  tree-structured reward functions.
\newblock In \emph{Proceedings of the 21st International Conference on
  Autonomous Agents and Multiagent Systems}, pp.\  118--126, 2022.

\bibitem[Bewley et~al.(2022)Bewley, Lawry, and Richards]{bewley2022summarising}
Tom Bewley, Jonathan Lawry, and Arthur Richards.
\newblock Summarising and comparing agent dynamics with contrastive
  spatiotemporal abstraction.
\newblock In \emph{IJCAI/ECAI Workshop on Explainable Artificial Intelligence},
  2022.

\bibitem[Bradley \& Terry(1952)Bradley and Terry]{bradley1952rank}
Ralph~Allan Bradley and Milton~E Terry.
\newblock Rank analysis of incomplete block designs: I. the method of paired
  comparisons.
\newblock \emph{Biometrika}, 39\penalty0 (3/4):\penalty0 324--345, 1952.

\bibitem[Breiman et~al.(2017)Breiman, Friedman, Olshen, and
  Stone]{breiman2017classification}
Leo Breiman, Jerome~H Friedman, Richard~A Olshen, and Charles~J Stone.
\newblock \emph{Classification and regression trees}.
\newblock Routledge, 2017.

\bibitem[Cao et~al.(2021)Cao, Wong, and Lin]{cao2021weak}
Zehong Cao, KaiChiu Wong, and Chin-Teng Lin.
\newblock Weak human preference supervision for deep reinforcement learning.
\newblock \emph{IEEE Transactions on Neural Networks and Learning Systems},
  32\penalty0 (12):\penalty0 5369--5378, 2021.

\bibitem[Christiano et~al.(2017)Christiano, Leike, Brown, Martic, Legg, and
  Amodei]{christiano2017deep}
Paul~F Christiano, Jan Leike, Tom Brown, Miljan Martic, Shane Legg, and Dario
  Amodei.
\newblock Deep reinforcement learning from human preferences.
\newblock \emph{Advances in neural information processing systems}, 30, 2017.

\bibitem[Chua et~al.(2018)Chua, Calandra, McAllister, and Levine]{chua2018deep}
Kurtland Chua, Roberto Calandra, Rowan McAllister, and Sergey Levine.
\newblock Deep reinforcement learning in a handful of trials using
  probabilistic dynamics models.
\newblock \emph{Advances in Neural Information Processing Systems}, 31, 2018.

\bibitem[Cobo et~al.(2012)Cobo, Isbell~Jr, and Thomaz]{cobo2012automatic}
Luis~C Cobo, Charles~L Isbell~Jr, and Andrea~L Thomaz.
\newblock Automatic task decomposition and state abstraction from
  demonstration.
\newblock Georgia Institute of Technology, 2012.

\bibitem[Coppens et~al.(2019)Coppens, Efthymiadis, Lenaerts, and
  Now{\'e}]{coppens2019distilling}
Youri Coppens, Kyriakos Efthymiadis, Tom Lenaerts, and Ann Now{\'e}.
\newblock Distilling deep reinforcement learning policies in soft decision
  trees.
\newblock In \emph{IJCAI/ECAI Workshop on Explainable Artificial Intelligence},
  2019.

\bibitem[Dao et~al.(2018)Dao, Mishra, and Lee]{dao2018deep}
Giang Dao, Indrajeet Mishra, and Minwoo Lee.
\newblock Deep reinforcement learning monitor for snapshot recording.
\newblock In \emph{2018 17th IEEE International Conference on Machine Learning
  and Applications (ICMLA)}, pp.\  591--598. IEEE, 2018.

\bibitem[De~Leeuw \& Mair(2009)De~Leeuw and Mair]{de2009multidimensional}
Jan De~Leeuw and Patrick Mair.
\newblock Multidimensional scaling using majorization: Smacof in r.
\newblock \emph{Journal of statistical software}, 31:\penalty0 1--30, 2009.

\bibitem[Devidze et~al.(2021)Devidze, Radanovic, Kamalaruban, and
  Singla]{devidze2021explicable}
Rati Devidze, Goran Radanovic, Parameswaran Kamalaruban, and Adish Singla.
\newblock Explicable reward design for reinforcement learning agents.
\newblock \emph{Advances in Neural Information Processing Systems},
  34:\penalty0 20118--20131, 2021.

\bibitem[Eric et~al.(2007)Eric, Freitas, and Ghosh]{eric2007active}
Brochu Eric, Nando Freitas, and Abhijeet Ghosh.
\newblock {Active Preference Learning with Discrete Choice Data}.
\newblock \emph{{Advances in Neural Information Processing Systems}}, 20, 2007.

\bibitem[Glanois et~al.(2021)Glanois, Weng, Zimmer, Li, Yang, Hao, and
  Liu]{glanois2021survey}
Claire Glanois, Paul Weng, Matthieu Zimmer, Dong Li, Tianpei Yang, Jianye Hao,
  and Wulong Liu.
\newblock A survey on interpretable reinforcement learning.
\newblock \emph{arXiv preprint arXiv:2112.13112}, 2021.

\bibitem[Gleave et~al.(2021)Gleave, Dennis, Legg, Russell, and
  Leike]{gleave2021quantifying}
Adam Gleave, Michael~D Dennis, Shane Legg, Stuart Russell, and Jan Leike.
\newblock Quantifying differences in reward functions.
\newblock In \emph{International Conference on Learning Representations}, 2021.
\newblock URL \url{https://openreview.net/forum?id=LwEQnp6CYev}.

\bibitem[Griffith et~al.(2013)Griffith, Subramanian, Scholz, Isbell, and
  Thomaz]{griffith2013policy}
Shane Griffith, Kaushik Subramanian, Jonathan Scholz, Charles~L Isbell, and
  Andrea~L Thomaz.
\newblock Policy shaping: Integrating human feedback with reinforcement
  learning.
\newblock \emph{Advances in neural information processing systems}, 26, 2013.

\bibitem[Gulliksen(1956)]{gulliksen1956least}
Harold Gulliksen.
\newblock A least squares solution for paired comparisons with incomplete data.
\newblock \emph{Psychometrika}, 21\penalty0 (2):\penalty0 125--134, 1956.

\bibitem[Guo et~al.(2018)Guo, Tian, Kalpathy-Cramer, Ostmo, Campbell, Chiang,
  Erdogmus, Dy, and Ioannidis]{guo2018experimental}
Yuan Guo, Peng Tian, Jayashree Kalpathy-Cramer, Susan Ostmo, J~Peter Campbell,
  Michael~F Chiang, Deniz Erdogmus, Jennifer~G Dy, and Stratis Ioannidis.
\newblock Experimental design under the bradley-terry model.
\newblock In \emph{IJCAI}, pp.\  2198--2204, 2018.

\bibitem[Haarnoja et~al.(2018)Haarnoja, Zhou, Abbeel, and
  Levine]{haarnoja2018soft}
Tuomas Haarnoja, Aurick Zhou, Pieter Abbeel, and Sergey Levine.
\newblock Soft actor-critic: Off-policy maximum entropy deep reinforcement
  learning with a stochastic actor.
\newblock In \emph{International Conference on Machine Learning}, pp.\
  1861--1870. PMLR, 2018.

\bibitem[Heuillet et~al.(2021)Heuillet, Couthouis, and
  D{\'\i}az-Rodr{\'\i}guez]{heuillet2021explainability}
Alexandre Heuillet, Fabien Couthouis, and Natalia D{\'\i}az-Rodr{\'\i}guez.
\newblock Explainability in deep reinforcement learning.
\newblock \emph{Knowledge-Based Systems}, 214:\penalty0 106685, 2021.

\bibitem[Hoffmann \& Magazzeni(2019)Hoffmann and
  Magazzeni]{hoffmann2019explainable}
J{\"o}rg Hoffmann and Daniele Magazzeni.
\newblock Explainable ai planning (xaip): overview and the case of contrastive
  explanation.
\newblock \emph{Reasoning Web. Explainable Artificial Intelligence}, pp.\
  277--282, 2019.

\bibitem[Huang et~al.(2018)Huang, Bhatia, Abbeel, and
  Dragan]{huang2018establishing}
Sandy~H Huang, Kush Bhatia, Pieter Abbeel, and Anca~D Dragan.
\newblock Establishing appropriate trust via critical states.
\newblock In \emph{2018 IEEE/RSJ International Conference on Intelligent Robots
  and Systems (IROS)}, pp.\  3929--3936. IEEE, 2018.

\bibitem[Huber et~al.(2019)Huber, Schiller, and Andr{\'e}]{huber2019enhancing}
Tobias Huber, Dominik Schiller, and Elisabeth Andr{\'e}.
\newblock Enhancing explainability of deep reinforcement learning through
  selective layer-wise relevance propagation.
\newblock In \emph{Joint German/Austrian Conference on Artificial Intelligence
  (K{\"u}nstliche Intelligenz)}, pp.\  188--202. Springer, 2019.

\bibitem[Jenner \& Gleave(2022)Jenner and Gleave]{jenner2022preprocessing}
Erik Jenner and Adam Gleave.
\newblock Preprocessing reward functions for interpretability.
\newblock \emph{arXiv preprint arXiv:2203.13553}, 2022.

\bibitem[Jiang et~al.(2019)Jiang, Hwang, and Lin]{jiang2019experience}
Wei-Cheng Jiang, Kao-Shing Hwang, and Jin-Ling Lin.
\newblock An experience replay method based on tree structure for reinforcement
  learning.
\newblock \emph{IEEE Transactions on Emerging Topics in Computing}, 9\penalty0
  (2):\penalty0 972--982, 2019.

\bibitem[Juozapaitis et~al.(2019)Juozapaitis, Koul, Fern, Erwig, and
  Doshi-Velez]{juozapaitis2019explainable}
Zoe Juozapaitis, Anurag Koul, Alan Fern, Martin Erwig, and Finale Doshi-Velez.
\newblock Explainable reinforcement learning via reward decomposition.
\newblock In \emph{IJCAI/ECAI Workshop on Explainable Artificial Intelligence},
  2019.

\bibitem[Kendall(1938)]{kendall1938new}
Maurice~G Kendall.
\newblock A new measure of rank correlation.
\newblock \emph{Biometrika}, 30\penalty0 (1/2):\penalty0 81--93, 1938.

\bibitem[Kendall(1975)]{kendall1975kendall}
MG~Kendall.
\newblock {Rank Correlation Methods; Griffin, C., Ed}, 1975.

\bibitem[Kingma \& Ba(2014)Kingma and Ba]{kingma2014adam}
Diederik~P Kingma and Jimmy Ba.
\newblock Adam: A method for stochastic optimization.
\newblock \emph{arXiv preprint arXiv:1412.6980}, 2014.

\bibitem[Knox \& Stone(2008)Knox and Stone]{knox2008tamer}
W~Bradley Knox and Peter Stone.
\newblock Tamer: Training an agent manually via evaluative reinforcement.
\newblock In \emph{2008 7th IEEE international conference on development and
  learning}, pp.\  292--297. IEEE, 2008.

\bibitem[Lafond et~al.(2013)Lafond, Tremblay, and Banbury]{lafond2013cognitive}
Daniel Lafond, S{\'e}bastien Tremblay, and Simon Banbury.
\newblock Cognitive shadow: A policy capturing tool to support naturalistic
  decision making.
\newblock In \emph{2013 IEEE International Multi-Disciplinary Conference on
  Cognitive Methods in Situation Awareness and Decision Support (CogSIMA)},
  pp.\  139--142. IEEE, 2013.

\bibitem[Lee et~al.(2021{\natexlab{a}})Lee, Smith, Dragan, and
  Abbeel]{lee2021bpref}
Kimin Lee, Laura Smith, Anca Dragan, and Pieter Abbeel.
\newblock B-pref: Benchmarking preference-based reinforcement learning.
\newblock In \emph{Thirty-fifth Conference on Neural Information Processing
  Systems Datasets and Benchmarks Track (Round 1)}, 2021{\natexlab{a}}.
\newblock URL \url{https://openreview.net/forum?id=ps95-mkHF_}.

\bibitem[Lee et~al.(2021{\natexlab{b}})Lee, Smith, and Abbeel]{lee2021pebble}
Kimin Lee, Laura~M Smith, and Pieter Abbeel.
\newblock Pebble: Feedback-efficient interactive reinforcement learning via
  relabeling experience and unsupervised pre-training.
\newblock In \emph{International Conference on Machine Learning}, pp.\
  6152--6163. PMLR, 2021{\natexlab{b}}.

\bibitem[Leike et~al.(2018)Leike, Krueger, Everitt, Martic, Maini, and
  Legg]{leike2018scalable}
Jan Leike, David Krueger, Tom Everitt, Miljan Martic, Vishal Maini, and Shane
  Legg.
\newblock Scalable agent alignment via reward modeling: a research direction.
\newblock \emph{arXiv preprint arXiv:1811.07871}, 2018.

\bibitem[Lindner et~al.(2021)Lindner, Turchetta, Tschiatschek, Ciosek, and
  Krause]{lindner2021information}
David Lindner, Matteo Turchetta, Sebastian Tschiatschek, Kamil Ciosek, and
  Andreas Krause.
\newblock Information directed reward learning for reinforcement learning.
\newblock \emph{Advances in Neural Information Processing Systems},
  34:\penalty0 3850--3862, 2021.

\bibitem[Liu et~al.(2018)Liu, Schulte, Zhu, and Li]{liu2018toward}
Guiliang Liu, Oliver Schulte, Wang Zhu, and Qingcan Li.
\newblock Toward interpretable deep reinforcement learning with linear model
  u-trees.
\newblock In \emph{Joint European Conference on Machine Learning and Knowledge
  Discovery in Databases}, pp.\  414--429. Springer, 2018.

\bibitem[Michaud et~al.(2020)Michaud, Gleave, and
  Russell]{michaud2020understanding}
Eric~J Michaud, Adam Gleave, and Stuart Russell.
\newblock Understanding learned reward functions.
\newblock \emph{arXiv preprint arXiv:2012.05862}, 2020.

\bibitem[Ng et~al.(2000)Ng, Russell, et~al.]{ng2000algorithms}
Andrew~Y Ng, Stuart Russell, et~al.
\newblock Algorithms for inverse reinforcement learning.
\newblock In \emph{Icml}, volume~1, pp.\ ~2, 2000.

\bibitem[Pan et~al.(2022)Pan, Bhatia, and Steinhardt]{pan2022the}
Alexander Pan, Kush Bhatia, and Jacob Steinhardt.
\newblock The effects of reward misspecification: Mapping and mitigating
  misaligned models.
\newblock In \emph{International Conference on Learning Representations}, 2022.
\newblock URL \url{https://openreview.net/forum?id=JYtwGwIL7ye}.

\bibitem[Puiutta \& Veith(2020)Puiutta and Veith]{puiutta2020explainable}
Erika Puiutta and Eric Veith.
\newblock Explainable reinforcement learning: A survey.
\newblock In \emph{International cross-domain conference for machine learning
  and knowledge extraction}, pp.\  77--95. Springer, 2020.

\bibitem[Rahtz et~al.(2022)Rahtz, Varma, Kumar, Kenton, Legg, and
  Leike]{rahtz2022safe}
Matthew Rahtz, Vikrant Varma, Ramana Kumar, Zachary Kenton, Shane Legg, and Jan
  Leike.
\newblock Safe deep rl in 3d environments using human feedback.
\newblock \emph{arXiv preprint arXiv:2201.08102}, 2022.

\bibitem[Reddy et~al.(2020)Reddy, Dragan, Levine, Legg, and
  Leike]{reddy2020learning}
Siddharth Reddy, Anca Dragan, Sergey Levine, Shane Legg, and Jan Leike.
\newblock Learning human objectives by evaluating hypothetical behavior.
\newblock In \emph{International Conference on Machine Learning}, pp.\
  8020--8029. PMLR, 2020.

\bibitem[Roth et~al.(2019)Roth, Topin, Jamshidi, and
  Veloso]{roth2019conservative}
Aaron~M Roth, Nicholay Topin, Pooyan Jamshidi, and Manuela Veloso.
\newblock Conservative q-improvement: Reinforcement learning for an
  interpretable decision-tree policy.
\newblock \emph{arXiv preprint arXiv:1907.01180}, 2019.

\bibitem[Rudin(2019)]{rudin2019stop}
Cynthia Rudin.
\newblock Stop explaining black box machine learning models for high stakes
  decisions and use interpretable models instead.
\newblock \emph{Nature Machine Intelligence}, 1\penalty0 (5):\penalty0
  206--215, 2019.

\bibitem[Russell \& Santos(2019)Russell and Santos]{russell2019explaining}
Jacob Russell and Eugene Santos.
\newblock {Explaining reward functions in Markov decision processes}.
\newblock In \emph{The Thirty-Second International Flairs Conference}, 2019.

\bibitem[Russell(2019)]{russell2019human}
Stuart Russell.
\newblock \emph{Human compatible: Artificial intelligence and the problem of
  control}.
\newblock Penguin, 2019.

\bibitem[Sadigh et~al.(2017)Sadigh, Dragan, Sastry, and
  Seshia]{sadigh2017active}
Dorsa Sadigh, Anca~D Dragan, Shankar Sastry, and Sanjit~A Seshia.
\newblock Active preference-based learning of reward functions.
\newblock In \emph{Proceedings of Robotics: Science and Systems (RSS)}, 2017.

\bibitem[Sanneman \& Shah(2022)Sanneman and Shah]{sanneman2022empirical}
Lindsay Sanneman and Julie~A Shah.
\newblock An empirical study of reward explanations with human-robot
  interaction applications.
\newblock \emph{IEEE Robotics and Automation Letters}, 2022.

\bibitem[Silva et~al.(2020)Silva, Gombolay, Killian, Jimenez, and
  Son]{pmlr-v108-silva20a}
Andrew Silva, Matthew Gombolay, Taylor Killian, Ivan Jimenez, and Sung-Hyun
  Son.
\newblock Optimization methods for interpretable differentiable decision trees
  applied to reinforcement learning.
\newblock In Silvia Chiappa and Roberto Calandra (eds.), \emph{Proceedings of
  the Twenty Third International Conference on Artificial Intelligence and
  Statistics}, volume 108 of \emph{Proceedings of Machine Learning Research},
  pp.\  1855--1865. PMLR, 26--28 Aug 2020.
\newblock URL \url{https://proceedings.mlr.press/v108/silva20a.html}.

\bibitem[Sternberg \& Horvath(1999)Sternberg and Horvath]{sternberg1999tacit}
Robert~J Sternberg and Joseph~A Horvath.
\newblock \emph{Tacit knowledge in professional practice: Researcher and
  practitioner perspectives}.
\newblock Psychology Press, 1999.

\bibitem[Tambwekar et~al.(2021)Tambwekar, Silva, Gopalan, and
  Gombolay]{tambwekar2021specifying}
Pradyumna Tambwekar, Andrew Silva, Nakul Gopalan, and Matthew Gombolay.
\newblock Specifying and interpreting reinforcement learning policies through
  simulatable machine learning.
\newblock \emph{arXiv preprint arXiv:2101.07140}, 2021.

\bibitem[Tien et~al.(2022)Tien, He, Erickson, Dragan, and Brown]{tien2022study}
Jeremy Tien, Jerry Zhi-Yang He, Zackory Erickson, Anca~D Dragan, and Daniel
  Brown.
\newblock A study of causal confusion in preference-based reward learning.
\newblock \emph{arXiv preprint arXiv:2204.06601}, 2022.

\bibitem[van~der Waa et~al.(2018)van~der Waa, van Diggelen, Bosch, and
  Neerincx]{van2018contrastive}
Jasper van~der Waa, Jurriaan van Diggelen, Karel van~den Bosch, and Mark
  Neerincx.
\newblock Contrastive explanations for reinforcement learning in terms of
  expected consequences.
\newblock In \emph{IJCAI/ECAI Workshop on Explainable Artificial Intelligence},
  2018.

\bibitem[Verma et~al.(2018)Verma, Murali, Singh, Kohli, and
  Chaudhuri]{verma2018programmatically}
Abhinav Verma, Vijayaraghavan Murali, Rishabh Singh, Pushmeet Kohli, and Swarat
  Chaudhuri.
\newblock Programmatically interpretable reinforcement learning.
\newblock In \emph{International Conference on Machine Learning}, pp.\
  5045--5054. PMLR, 2018.

\bibitem[Wilde et~al.(2020)Wilde, Blidaru, Smith, and
  Kuli{\'c}]{wilde2020improving}
Nils Wilde, Alexandru Blidaru, Stephen~L Smith, and Dana Kuli{\'c}.
\newblock Improving user specifications for robot behavior through active
  preference learning: Framework and evaluation.
\newblock \emph{The International Journal of Robotics Research}, 39\penalty0
  (6):\penalty0 651--667, 2020.

\bibitem[Wirth et~al.(2016)Wirth, F{\"u}rnkranz, and Neumann]{wirth2016model}
Christian Wirth, Johannes F{\"u}rnkranz, and Gerhard Neumann.
\newblock Model-free preference-based reinforcement learning.
\newblock In \emph{Thirtieth AAAI Conference on Artificial Intelligence}, 2016.

\bibitem[Zahavy et~al.(2016)Zahavy, Ben-Zrihem, and Mannor]{zahavy2016graying}
Tom Zahavy, Nir Ben-Zrihem, and Shie Mannor.
\newblock Graying the black box: Understanding dqns.
\newblock In \emph{International conference on machine learning}, pp.\
  1899--1908. PMLR, 2016.

\bibitem[Zhu et~al.(2018)Zhu, Huang, and Zhang]{zhu2018object}
Guangxiang Zhu, Zhiao Huang, and Chongjie Zhang.
\newblock Object-oriented dynamics predictor.
\newblock \emph{Advances in Neural Information Processing Systems}, 31, 2018.

\end{thebibliography}
\bibliographystyle{iclr2023_conference}

\appendix

\renewcommand{\thefigure}{A\arabic{figure}}
\setcounter{figure}{0}

\newpage
\section{Methodological Details}
\label{app:meth_details}

\subsection{Sign Constraint for Return Estimates}
\label{app:sign_constraint}

Applying a sign constraint to the trajectory-level return estimates means that rewards output by a reward tree (via Equation \ref{eq:reward_prediction}) are also all either positive or negative. This has no effect on any measure of preference reconstruction since preferences are invariant to affine transformations of an underlying utility function. However, we find it brings two distinct benefits:
\begin{itemize}
    \item Enabling the prevention of perverse incentives for agents to terminate or elongate episodes in tasks with termination conditions (negative rewards on non-terminal transitions incentivise termination, while positive rewards incentivise elongation). 
    \item Simplifying the manual interpretation of tradeoffs between rewards from different leaves of a tree (understanding the relative impacts of ``more of a negative reward" and ``less of a positive reward" requires the awkward mental juggling of negatives).
\end{itemize}
For the task with a termination condition in this paper (Land), we use negative rewards ($\text{max}=0$ constraint) to disincentivise episode elongation, because termination is generally indicative of success. For the two fixed-length tasks (Follow and Chase) we default to using positive rewards ($\text{min}=0$ constraint). Although this is arbitrary, our own experience is that positive rewards make for somewhat more intuitive interpretation of the tree structure, and its effect on agent actions. We stress that this is purely anecdotal; the relative human interpretability of positive, negative and mixed-sign rewards would be a worthy subject for deeper empirical investigation.

\subsection{Pseudocode for Online Algorithm}
\label{app:pseudocode}

See Algorithm \ref{alg:online_algorithm} below.

\newpage
\begin{algorithm}[H]
\caption{Online preference-based reward tree learning}\label{alg:online_algorithm}
\begin{algorithmic}[1]
\Statex \textbf{Inputs:} Possibly pre-trained dynamics model $D'$, feature function $\phi$, trajectory budget $N_\text{max}$,
\Statex preference budget $K_\text{max}$, tree size limit $L_\text{max}$, tree size regularisation $\alpha$
\vspace{0.2cm}
\State Initialise empty preference graph $\Xi \gets \emptyset$, $\mathcal{L}\gets\emptyset$
\State Initialise one-leaf tree $\mathcal{T}$ with $\textbf{r}\gets[0]$
\For{$i\in\{1,...,N_\text{max}\}$}
    \State Initialise time $t\gets 0$ and environment state $s^i_0$
    \While{episode not yet terminated} \Comment{Model-based trajectory generation (Sec.\ref{sec:model_based})}
        \State Compute action $a^i_t$ using PETS algorithm
        \Statex \hspace{1cm} with $D'$ and rewards via Equation \ref{eq:reward_prediction}
        \State Send $a^i_t$ to environment and get next state $s^i_{t+1}$
        \State Update $D'$ on recent transitions \Comment{May not be required; see Appendix \ref{app:pets_details}}
        \State $\textbf{x}_{t+1}^i\gets\phi(s_t^i,a_t^i,s_{t+1}^i)$
        \State $t\gets t+1$
    \EndWhile
    \State $\xi^i\gets(\textbf{x}_1^i,...,\textbf{x}_{T^i}^i)$
    \State $K_\text{batch} \gets \text{min}((K_\text{max}-|\mathcal{L}|)/(N_\text{max}+1-i),|\Xi|)$
    \For{$k\in\{1,...,K_\text{batch}\}$} \Comment{Preference batch collection (Sec \ref{sec:pbrl})}
        \State Sample $\xi^j$ from $\Xi$ uniformly without replacement
        \State Query human for preference $\xi^i\succ\xi^j$ or $\xi^j\succ\xi^i$
        \State $\mathcal{L}\gets\mathcal{L}\cup\left\{\begin{array}{ll}\{(i,j)\}&\text{if }\xi^j\succ\xi^i\\\{(j,i)\}&\text{otherwise}\end{array}\right.$
    \EndFor
    \State $\Xi\gets\Xi\cup\{\xi^i\}$
    \If{$|\mathcal{L}|>0$}
        \State Compute $\textbf{g}$ via Equation \ref{eq:proxy_loss_trajectory_level} \Comment{Trajectory-level return estimation (Sec \ref{sec:traj_level})}
        \State Initialise one-leaf tree $\mathcal{T}$
        \State $\mathcal{C}\gets$ midpoints between per-feature unique values in $\Xi$
        \While{$L_\mathcal{T}<L_\text{max}$} \Comment{Tree growth (Section \ref{sec:growth})}
            \For{$l\in\{1,...,L_\mathcal{T}\}$}
                \For{$f\in\{1,...,F\}$}
                    \For{$c\in\mathcal{C}_f$}
                        \State Compute $\ell_{\text{0-1}}$ reduction for $\mathcal{T}+[lfc]$ via Equation  \ref{eq:split_criterion}
                    \EndFor
                \EndFor
            \EndFor
            \If{$\text{max}(\ell_{\text{0-1}}\text{ reduction})\leq 0$}
                \State \textbf{break} \Comment{Stop tree growth early}
            \EndIf
            \State $l,f,c\gets\text{argmax}(\ell_{\text{0-1}}\text{ reduction})$
            \State $\mathcal{T}\gets\mathcal{T}+[lfc]$
        \EndWhile
        \State $\mathbb{T}=()$
        \While{$L_\mathcal{T}>1$} \Comment{Tree pruning (Section \ref{sec:pruning})}
            \For{$l\in\{1,...,L_\mathcal{T}\}$}
                \State Compute $\ell_{\text{0-1}}$ reduction for $\mathcal{T}-[l]$ \Comment{$\mathcal{T}-[l]$ denotes pruning $l$th leaf}
            \EndFor
            \State $l\gets\text{argmax}(\ell_{\text{0-1}}\text{ reduction})$
            \State $\mathcal{T}\gets\mathcal{T}-[l]$
            \State Append $\mathcal{T}$ to $\mathbb{T}$
        \EndWhile
        \State $\mathcal{T}\leftarrow\text{argmin}_{\mathcal{T}\in\mathbb{T}}(\ell_{\text{0-1}}\text{ plus }\alpha\text{-scaled tree size})$
    \EndIf
\EndFor
\end{algorithmic}
\end{algorithm}



\newpage
\section{Comparison to Model-free Reinforcement Learning}
\label{app:sac}

\begin{figure}[H]
\centering
\includegraphics[width=1\textwidth]{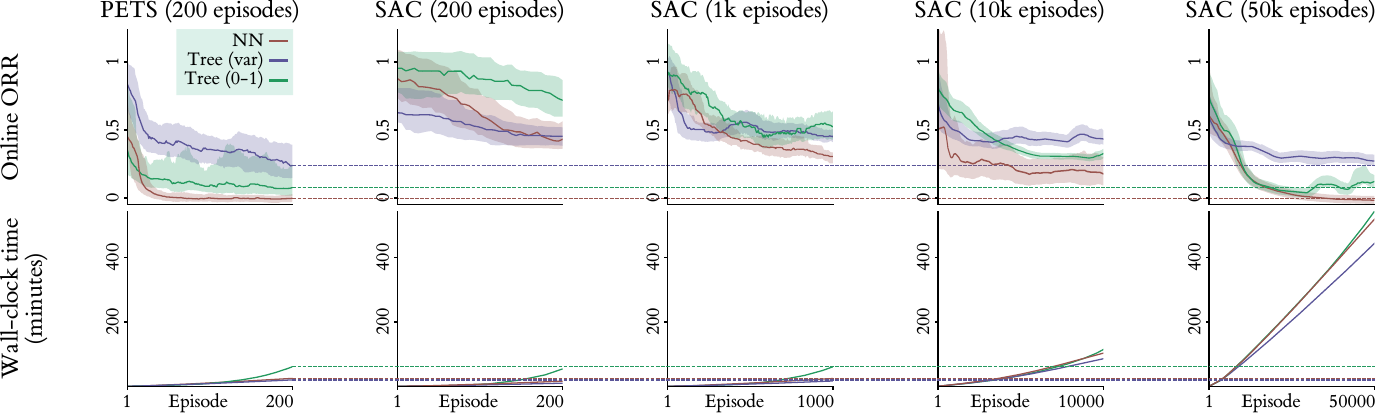}
\caption{Comparing the use of PETS (model-based) and SAC (model-free) agents on the Follow task (see Appendix \ref{app:tasks_and_oracles} for task details). The PETS results are taken directly from Figure \ref{fig:results_main}. For SAC we retain the total preference budget of $K_\text{max}=1000$, but for longer runs add episode trajectories to $\Xi$ and $\mathcal{L}$ at a reduced frequency so that $N_\text{max}=200$ (e.g. for $50000$ episodes, only $1$ in $250$ episodes are added to the graph; the rest are skipped). All SAC agents use policy and value networks with two $256$-unit hidden layers each, learning rates of $1e^{-4}$ and $1e^{-3}$ for the policy and value updates with the Adam optimiser, a discount factor of $\gamma=0.99$, and an interpolation factor of $0.99$ for Polyak averaging of the target networks. Updates use mini-batches of $32$ transitions sampled uniformly from a rolling replay buffer of capacity $5e^{4}$.\\
We find that initially running SAC for a total of $200$ episodes, matching our PETS experiments, gives the model-free learning algorithm insufficient time to achieve good performance in terms of regret on the oracle reward function
(a higher learning rate leads SAC to become unstable). We then progressively increase the length of SAC runs until regret performance matches the use of PETS, and find that this requires around $50000$ episodes, an increase of $250$ times. It is noteworthy that the variance-based tree model seems to perform best in the short-runtime regime, but worst in the long-runtime regime. Time constraints prevented us from investigating whether this holds in other tasks, but such an investigation would be worthwhile.\\
In terms of wall-clock time (on a single NVIDIA Tesla P100 GPU), running reward learning with SAC for $1000$ episodes is roughly equivalent to $200$ episodes using PETS ($25$-$60$ minutes, depending on the reward model architecture). For $50000$ episodes, this time increases to $9$ hours. This brings a very practical disadvantage: if reward learning were done using human preferences instead of an oracle, that person would have to dedicate more than a full working day to the exercise, most of which would be spent waiting for several minutes between each successive preference batch.
}
\label{fig:sac_appendix}
\end{figure}

One of the most consistently observed benefits of model-based RL is its sample efficiency, and this trend holds in our context. Running Algorithm \ref{alg:online_algorithm} unchanged except for the use of a soft actor-critic (SAC) agent for policy learning, we find that approximately two orders of magnitude more environment interaction is required to achieve equivalent performance in terms of regret at convergence. In turn, this increases wall-clock runtime by $10$-$20$ times, thereby outweighing the higher per-timestep computational cost of PETS over SAC. The caption of Figure \ref{fig:sac_appendix} gives further details.

\newpage
\section{Aircraft Handling Environment}
\label{app:env_details}

\subsection{Motivation}

Pilots of fast jet aircraft require exceptional handling abilities, acquired over years of advanced training.
There would be immense practical value in developing a method for distilling the knowledge and preferences of pilots and other domain experts into a software model that captures realistic handling behaviour. The scalability of such a model would make it useful for strategic planning exercises, training of a range of operational roles, and development and testing of other software systems. However, as in many contexts where intuitive decision-making and rapid motor control are paramount,
the preferences of experts
(over the space of fast jet handling trajectories) are in large part tacit, and thus defy direct scrutiny or verbal description. Put simply:
experts
know good handling when they see it, but cannot directly express \textit{why}.\footnote{This statement certainly underestimates the rich complexity of
human
expertise; in reality, an
expert's
mental model is likely to be partly tacit and partly explicit. The general strategy of preference-based reward learning is to operate \textit{as if} the mental model were $100\%$ tacit, and explore what can be achieved under such a strong restriction. Real-world applications would likely benefit from combining this approach with some amount of hand-coded expert knowledge.} This makes it practically challenging to accurately elicit
this
knowledge for codification into an automated system.

The methods presented in this paper form the basis of a possible solution to this dilemma. Given a dataset of trajectories executed by an artificial learning agent and labelled with pairwise
expert
preferences (which require only tacit knowledge to produce), we use statistical learning algorithms to construct an interpretable explanatory model of those preferences. The result is two distinct outputs that could form valuable components of future
planning, training and development
software:
\begin{enumerate}
    \item A tree-structured reward function, which may be used for automated scoring of flight trajectories executed by
    human or artificial pilots.
    We aim for this to produce an evaluation that is consistent, unbiased and aligned with the judgement that
    the original expert
    would have made, alongside an explanatory rationale that
    can be leveraged to justify, verify and improve handling behaviour.
    \item A model-based RL agent capable of
    executing
    high-quality handling behaviour with respect to the reward function,
    for use in simulation.
\end{enumerate}

It should be noted that 
any realistic handling scenario would involve multiple experts
somewhat-differing knowledge and expertise. A natural extension of our approach, which we see as valuable future work, is to learn individual reward functions for each
expert,
then leverage the intrinsic interpretability to identify biases, inconsistencies and tradeoffs. This suggests a third application of reward tree learning: providing a basis for evaluating and training the
experts themselves.

\subsection{Implementation}
\label{app:env_implementation}

We consider a simple set-piece formulation of the aircraft handling problem, in which the piloting agent is given a short time window to manoeuvre their aircraft (the ego jet, EJ) in a particular manner relative to a second reference jet (RJ). Special cases of this formulation create a wide variety of tasks for the pilot to solve. Options include:
\begin{itemize}
\item RJ is a \textit{friendly} aircraft which EJ should accompany in formation flight.
\item RJ is \textit{adversarial} and EJ must outmanoeuvre it to gain a tactical advantage.
\item Rather than being a distinct physical entity, RJ defines a \textit{goal pose} (position and attitude) for EJ to reach. The goal pose may be fixed or moving over time.
\end{itemize}
We developed this formulation to strike a balance between simplicity and generality; many realistic scenarios faced by a fast jet pilot involve interaction with a single other airborne entity. On a practical level, it provides scope for the definition of many alternative tasks given the same state and action spaces, and largely unchanged dynamics.

The state space contains the positions, attitudes, velocities and accelerations of both EJ and RJ (state dimensionality $=37$) and the action space consists of pitch, roll, yaw and thrust demands for EJ only (action dimensionality $=4$). The EJ dynamics function integrates these demands with a simplified physics engine, including gravity and air resistance (we make no claim of realism here; the simulator is merely a proof of concept). A new action is accepted every $25$ steps of the physics engine, reducing an agent's decision frequency to approximately $1$Hz. RJ dynamics, as well as the conditions of state initialisation and termination, vary between tasks (see Appendix \ref{app:tasks_and_oracles}). 

The final generic aspect of the implementation is the feature function $\phi$, which maps the transition space $\mathcal{S}\times\mathcal{A}\times\mathcal{S}$ (total dimensionality $=37+4+37=78$) into an $F$-dimensional space of task-relevant features. In consultation with engineers with experience of aerospace simulation and control algorithms, we devised the following set of $F=30$ features that is sufficiently expressive to capture the important information for all three of our target tasks, without being overly specialised to one or providing too much explicit guidance to the reward learning process. Apart from those containing ``\texttt{delta}" or ``\texttt{rate}", all features are computed over the successor state for each transition, $s_{t+1}$.

\begin{table}[H]
\centering
\small
\begin{tabular}{|c|c|}
\texttt{dist}                 & Euclidean distance between EJ and RJ \\
\texttt{closing}\ \texttt{speed}       & Closing speed between EJ and RJ (negative = moving closer) \\
\texttt{alt}                  & Altitude of EJ \\
\texttt{alt}\ \texttt{error}           & Difference in altitude between EJ and RJ (negative = EJ is lower) \\
\texttt{delta}\ \texttt{alt}\ \texttt{error}    & Change in \texttt{alt}\ \texttt{error} between $s_t$ and $s_{t+1}$ \\
\texttt{dist}\ \texttt{hor}             & Euclidean distance between EJ and RJ in horizontal plane \\
\texttt{delta}\ \texttt{dist}\ \texttt{hor}      &  Change in \texttt{dist}\ \texttt{hor} between $s_t$ and $s_{t+1}$ (negative = moving closer) \\
\texttt{pitch}\ \texttt{error}         & Absolute difference in pitch angle between EJ and RJ \\
\texttt{delta}\ \texttt{pitch}\ \texttt{error}  & Change in \texttt{pitch}\ \texttt{error} between $s_t$ and $s_{t+1}$ \\
\texttt{abs}\ \texttt{roll}            & Absolute roll angle of EJ \\
\texttt{roll}\ \texttt{error}          & Absolute difference in roll angle between EJ and RJ \\
\texttt{delta}\ \texttt{roll}\ \texttt{error}   & Change in \texttt{roll}\ \texttt{error} between $s_t$ and $s_{t+1}$ \\
\texttt{hdg}\ \texttt{error}           & Absolute difference in heading angle between EJ and RJ \\
\texttt{delta}\ \texttt{hdg}\ \texttt{error}    & Change in \texttt{hdg}\ \texttt{error} between $s_t$ and $s_{t+1}$ \\
\texttt{fwd}\ \texttt{error}           & Angle between 3D vectors indicating forward axes of EJ and RJ \\
\texttt{delta}\ \texttt{fwd}\ \texttt{error}    &  Change in \texttt{fwd}\ \texttt{error} between $s_t$ and $s_{t+1}$ \\
\texttt{up}\ \texttt{error}            & Angle between 3D vectors indicating upward axes of EJ and RJ \\
\texttt{delta}\ \texttt{up}\ \texttt{error}     & Change in \texttt{up}\ \texttt{error} between $s_t$ and $s_{t+1}$ \\
\texttt{right}\ \texttt{error}         & Angle between 3D vectors indicating rightward axes of EJ and RJ \\
\texttt{delta}\ \texttt{right}\ \texttt{error}  & Change in \texttt{right}\ \texttt{error} between $s_t$ and $s_{t+1}$ \\
\texttt{los}\ \texttt{error}           & Angle between forward axis of EJ and vector from EJ to RJ \\ & (measures whether RJ is in EJ's line of sight) \\
\texttt{delta}\ \texttt{los}\ \texttt{error}    & Change in \texttt{los}\ \texttt{error} between $s_t$ and $s_{t+1}$ \\
\texttt{abs}\ \texttt{lr}\ \texttt{offset}      & Magnitude of projection of vector from EJ to RJ onto RJ's rightward axis \\ & (measures left-right offset between the two aircraft in RJ's reference frame) \\
\texttt{speed}                & Airspeed of EJ \\
\texttt{g}\ \texttt{force}             & Instantaneous g-force experienced by EJ \\
\texttt{pitch}\ \texttt{rate}          & Absolute change of EJ pitch between $s_t$ and $s_{t+1}$ \\
\texttt{roll}\ \texttt{rate}           & Absolute change of EJ roll between $s_t$ and $s_{t+1}$ \\
\texttt{yaw}\ \texttt{rate}            & Absolute change of EJ yaw between $s_t$ and $s_{t+1}$ \\
\texttt{thrust}               & Instantaneous thrust output by EJ engines \\
\texttt{delta}\ \texttt{thrust}        & Absolute change in \texttt{thrust} between $s_t$ and $s_{t+1}$
\end{tabular}
\end{table}


\subsection{Tasks and Oracles}
\label{app:tasks_and_oracles}

In this paper, we consider three concrete tasks that instantiate the general EJ-RJ framework. For each, we construct a plausible oracle reward function from a subset of the $30$ features, meaning that reward learning is in part a feature selection problem (tree models perform feature selection explicitly whenever they add a new splitting rule). Although the precise nature of the oracle reward functions is secondary, and those given below are among many equally reasonable alternatives, we dedicated several hours of development time to ensuring they yield reasonable behaviour upon visual inspection. The difficulty and seeming arbitrariness of this manual reward design process is precisely why reward learning (ultimately from real human preferences) is an enticing proposition. Descriptions of the three tasks, along with their respective oracles, are given below:

\begin{itemize}
\item \textbf{Follow}: Here RJ follows a linear horizontal flight path at a constant velocity, which is oriented opposite to the initial velocity of EJ. The goal is to turn onto and then maintain the path up to the episode time limit of $20$ timesteps. This constitutes a very simple form of formation flight. The oracle reward function incentivises closing the distance to the moving target, and matching the upward axes of EJ and RJ:
$$
r=-(\texttt{dist}+0.05\times\texttt{closing}\ \texttt{speed}+10\times\texttt{up}\ \texttt{error}).
$$
\item \textbf{Chase}: Here RJ follows an erratic trajectory generated by random control inputs, and the goal is to chase it without taking EJ below a safe altitude of $50$. Episodes terminate after $20$ timesteps. The oracle reward function incentivises keeping RJ at a distance of $20$ and within EJ's line of sight, while keeping EJ oriented upright. It also has a large penalty for dropping below the safe altitude:
$$
r=-(\text{abs}(\texttt{dist}-20)+10\times\texttt{los}\ \texttt{error}+5\times\texttt{abs}\ \texttt{roll}+\left\{\begin{array}{ll}100&\text{if }\texttt{alt}<50\\0&\text{otherwise}\end{array}\right.).
$$
\item \textbf{Land}: Here the goal is to execute a safe approach towards landing on a runway, where RJ represents the ideal landing position (central, zero altitude, slight upward pitch). EJ is initialised at a random altitude, pitch, roll and offset, such that landing may be challenging but always physically possible. An episode terminates if EJ passes RJ along the axis of the runway, or after $25$ timesteps otherwise. The oracle reward function for this task is by far the most complex of the three, including terms that incentivise continual descent, penalise g-force and engine thrust, and punish the agent for making contact with the ground ($\texttt{alt}<0.6$) before the start of the runway:
\begin{align*}
r=-(0.05\times\texttt{abs}\ \texttt{lr}\ \texttt{offset}+0.05\times\texttt{alt}
+\texttt{hdg}\ \texttt{error}+\texttt{abs}\ \texttt{roll}\\+0.5\times\texttt{pitch}\ \texttt{error}+0.25\times(\texttt{yaw}\ \texttt{rate}+\texttt{roll}\ \texttt{rate}+\texttt{pitch}\ \texttt{rate})\\+0.1\times\texttt{g}\ \texttt{force}+0.025\times\texttt{thrust}+0.05\times\texttt{delta}\ \texttt{thrust}\\+\left\{\begin{array}{ll}1&\text{if }\texttt{delta}\ \texttt{dist}\ \texttt{hor}>0\\0&\text{otherwise}\end{array}\right.+\left\{\begin{array}{ll}2&\text{if }\texttt{delta}\ \texttt{alt}>0\\0&\text{otherwise}\end{array}\right.\\+\left\{\begin{array}{ll}1&\text{if }\texttt{abs}\ \texttt{lr}\ \texttt{offset}>10\\0&\text{otherwise}\end{array}\right.+\left\{\begin{array}{ll}10&\text{if }\texttt{alt}<0.6\\0&\text{otherwise}\end{array}\right.).
\end{align*}

\end{itemize}

\newpage
\section{Implementation and Experiment Details}
\label{app:exp_details}

\subsection{Oracle Preferences}

Oracle preferences are generated in accordance with the Bradley-Terry model given in Equation \ref{eq:bradley-terry}, i.e. by computing the returns for the two trajectories $\xi^i$ and $\xi^j$, and sampling from a Boltzmann distribution parameterised by those returns. In our main experiments, we set the temperature coefficient $\beta=0$, which results in the oracle deterministically selecting the trajectory with higher return (ties broken uniform-randomly). In Section \ref{sec:sensitivity_analysis} we study cases with $\beta>0$, which provide a more realistic emulation of real human preference data.

\subsection{Hyperparameters for Tree Induction}

In all experiments, we use the following hyperparameters during tree induction. These were identified through informal search, and we make no claim of optimality, but they do lead to reasonable performance on the three tasks of varying complexity. This indicates a general insensitivity of the method to precise hyperparameter values, which is often practically advantageous.
\begin{itemize}
    \item Trajectory return estimation using the Adam optimiser with a learning rate of $0.1$. Optimisation stops when the mean $\ell_{\text{NLL}}$ changes by $<1e^{-5}$ between successive gradient steps.
    \item Per-feature candidate split thresholds $\mathcal{C}$ defined as all midpoints between adjacent unique values in the trajectory set $\Xi$. These are recomputed on each update.
    \item Tree size limit $L_\text{max}=100$.
    \item Tree size regularisation coefficient $\alpha=5e^{-3}$.
\end{itemize}

As mentioned in Appendix \ref{app:sign_constraint}, we enforce negative rewards ($\text{max}=0$ constraint) for the Land task, and positive rewards ($\text{min}=0$ constraint) for Follow and Chase. 

\subsection{Model-based RL Implementation}
\label{app:pets_details}

For conceptual details on the PETS algorithm, we refer readers to the original paper by \citet{chua2018deep}. In our implementation, the dynamics model is an ensemble of five NNs, each with four hidden layers of $200$ hidden units and ReLU activations. State vectors are pre-normalised by applying a hand-specifed scale factor to each dimension. Decision-time planning operates over a time horizon of $H=10$ and consists of $10$ iterations of the cross-entropy method. Each iteration samples $20$ candidate action sequences from an independent Gaussian, of which the top $5$ in terms of return are identified as \textit{elites}, then updates the sampling Gaussian towards the elites with a learning rate of $0.5$. In all experiments we use $\gamma=1$, meaning no temporal discounting is applied during planning.

In our experiments, we find that the particular dynamics of the aircraft handling environment permit us to pre-train $D'$ on random offline data, and accurately generalise to states encountered during online reward learning. This means we perform no further updates to the model while reward learning is ongoing. As well as improving wall-clock speed, this avoids complexity and convergence issues arising from having two interacting learning processes (note that simultaneous learning is completely unavoidable with model-free RL). To pre-train, we collect $1e^5$ transitions by rolling out a uniform random policy, then update each of the five networks on $1e^5$ independently sampled mini-batches of $256$ transitions, using the mean squared error loss over normalised next-state predictions.

\subsection{Neural Network Reward Learning Baseline}
\label{app:nn_baseline}

We baseline our reward tree models against the de facto standard approach of reward learning using a NN. In constructing this baseline, we aimed to retain as much of Algorithm \ref{alg:online_algorithm} as possible, so that only the model architecture varies. The result is that we replace lines $21$-$47$ with the following:

\begin{algorithm}[H]
\begin{algorithmic}[1]
\setcounter{ALG@line}{20}
\For{$m\in\{1,...,M\}$}
    \State $\mathcal{L}_\text{mini-batch}\gets$ a mini-batch of $B$ preference labels sampled from $\mathcal{L}$
    \State Compute $\ell_{\text{NLL}}$ over $\mathcal{L}_\text{mini-batch}$ via Equations \ref{eq:bradley-terry} and \ref{eq:loss_functions}
    \State Backpropagate loss and update network parameters
\EndFor
\State $\textbf{r}_\text{all}\gets$ reward predictions for all feature vectors in $\Xi$
\State Scale network outputs by $1/\text{std}(\textbf{r}_\text{all})$
\State Shift network outputs by $-\text{min}(\textbf{r}_\text{all})$ or $-\text{max}(\textbf{r}_\text{all})$, depending on desired reward sign
\end{algorithmic}
\end{algorithm}

The new lines $26$-$28$ replicate the two constraints applied in Equation \ref{eq:proxy_loss_trajectory_level}.

In all experiments, we follow \citet{lee2021pebble} in implementing the reward model as a three-layer network with 256 hidden units each and leaky ReLU activations, and performing the update on line $24$ using the Adam optimiser \citep{kingma2014adam} with a learning rate of $3e^{-4}$. On each update, we sample $M=100$ mini-batches of size $B=32$ and take one gradient step per mini-batch.

\subsection{Coverage Datasets for Policy-Invariant Evaluation}
\label{app:coverage_datasets}

\citet{gleave2021quantifying} recently highlighted the importance of comparing and evaluating learnt reward functions in a policy-invariant manner, by using a common evaluation distribution rather than on-policy data generated by agents optimising for each reward. Ideally, the offline evaluation data should have high coverage (i.e. high-entropy state distribution, both high- and low-quality trajectories), in order to characterise the reward functions' outputs across a spectrum of plausible policies.

In our context, we can generate data that satisfies these requirements by leveraging the known oracle reward functions and the PETS algorithm. We deploy PETS using the oracle reward, but randomise the planning parameters (number of planning iterations $\in\{1,...,50\}$, number of action sequence samples $\in\{4,...,50\}$) on every episode. In all cases, we take the top $25\%$ of action sequences as elites. This randomisation results in trajectories that are sometimes near-optimal with respect to the oracle, sometimes moderate in quality, and sometimes barely better than random. For all three tasks, we generate a dataset of $200$ evaluation trajectories in this manner.

\newpage

\section{Visualing Reward Functions with Similarity Embeddings}
\label{app:rank_embeddings}

Here we briefly discuss a novel form of reward visualisation which we developed and found valuable during our own analysis. Given some measure of similarity between reward functions, such as the EPIC metric proposed by \citet{gleave2021quantifying}, we can compute a matrix of pairwise similarities between any number of such functions (computational cost permitting). We can then produce a 2D embedding of the functions by applying multidimensional scaling (MDS). Visualising this embedding as a scatter plot enables the discovery of salient patterns and trends in the set of functions. 

\begin{figure}[H]
\centering
\includegraphics[width=1\textwidth]{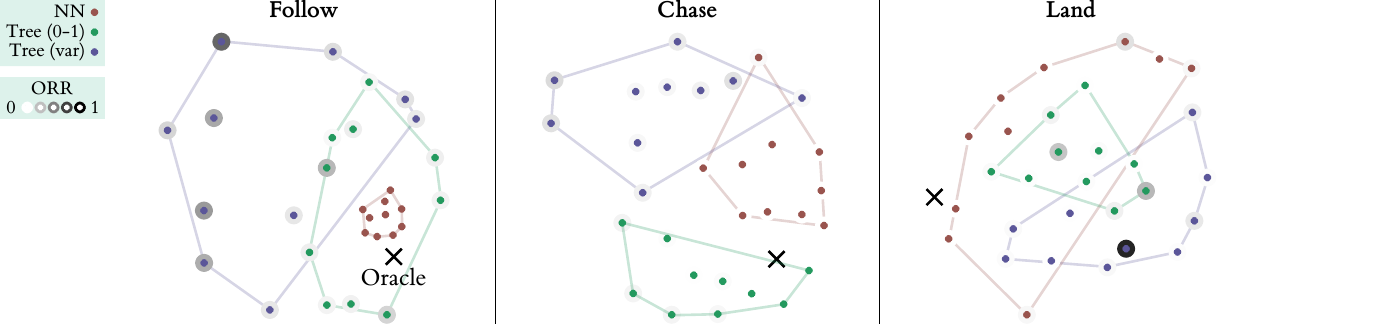}
\caption{Rank correlation embeddings for all model repeats from the main experiments, with the scatter point for each repeat shaded by ORR.
}
\label{fig:embeddings}
\end{figure}

In Figure \ref{fig:embeddings}, we use the metric of rank correlation on the coverage datasets, and the SMACOF MDS algorithm \citep{de2009multidimensional}, to embed all $30$ model repeats and the oracle for each task. This gives an impression of the models' similarity not just to the oracle, but to each other. Aside from the Follow NNs, which form a tight cluster near the oracle, the distribution for each model indicates roughly equal consistency between repeats. The overlap of convex hulls suggests that the rankings produced by all models are broadly similar for Land, but more distinct for Chase. Shading points by ORR reveals that while models further from the oracle tend to induce worse-performing policies, the trend is not monotonic. This reinforces the point made elsewhere that the problems of learning a good policy and exactly replicating the ground truth reward are not identical.

Populating such embedding plots more densely, perhaps by varying model hyperparameters, could provide a means of mapping the space of learnable reward functions and its relationship to policy performance. It would also be straightforward to compute similarity values for the same model repeat at multiple checkpoints during learning. This would yield a trajectory in the rank embedding space, which could aid the assessment of the stability and convergence properties of online learning with different models and hyperparameter values.

\end{document}